\documentclass[sn-mathphys, iicol]{sn-jnl}% Default with double column layout

\usepackage{graphicx}
\usepackage{multirow}
\usepackage{amsmath,amssymb,amsfonts}
\usepackage{amsthm}
\usepackage{mathrsfs}
\usepackage[title]{appendix}
\usepackage{xcolor}
\usepackage{textcomp}
\usepackage{manyfoot}
\usepackage{booktabs}
\usepackage{algorithm}
\usepackage{algorithmicx}
\usepackage{algpseudocode}
\usepackage{listings}
\usepackage{tabularx}
\usepackage{makecell}
\usepackage{balance}
\usepackage{times}
\usepackage{epsfig}
\usepackage{tikz}
\usepackage{comment}
\usepackage{bbding}
\usepackage{newfloat}
\usepackage{array}
\usepackage{stfloats}
\usepackage{url}
\usepackage{verbatim}

\theoremstyle{thmstyleone}%
\theoremstyle{thmstyletwo}%

\theoremstyle{thmstylethree}%

\begin{document}

\title[Article Title]{Rethinking Visual Prompt Learning as Masked Visual Token Modeling}

\author[1]{\fnm{Ning} \sur{Liao}}\email{liaoning@sjtu.edu.cn}

\author[1]{\fnm{Bowen} \sur{Shi}}\email{sjtu\_shibowen@sjtu.edu.cn}

\author[2]{\fnm{Xiaopeng} \sur{Zhang}}\email{zxphistory@gmail.com}

\author[3]{\fnm{Min} \sur{Cao}}\email{caomin0719@126.com}

\author*[1]{\fnm{Junchi} \sur{Yan}}\email{yanjunchi@sjtu.edu.cn}

\author[2]{\fnm{Qi} \sur{Tian}}\email{tian.qi1@huawei.com}

\affil[1]{\orgname{Shanghai Jiao Tong University}, \orgaddress{\city{Shanghai}, \postcode{200240}, \country{China}}}

\affil[2]{\orgdiv{Huawei Inc.}, \orgaddress{\city{Shenzhen}, \postcode{518129}, \country{China}}}

\affil[3]{\orgname{Soochow University}, \orgaddress{\city{Suzhou}, \postcode{215006}, \country{China}}}

\abstract{
Prompt learning has achieved great success in efficiently exploiting large-scale pre-trained models in natural language processing (NLP). It reformulates the downstream tasks as the generative pre-training ones to achieve consistency, thus improving the performance stably. However, when transferring it to the vision area, current visual prompt learning methods are almost designed on discriminative pre-trained models, and there is also a lack of careful design to unify the forms of pre-training and downstream tasks. To explore prompt learning on the generative pre-trained visual model, as well as keeping the task consistency, we propose Visual Prompt learning as masked visual Token Modeling (VPTM) to transform the downstream visual classification into the pre-trained masked visual token prediction. In addition, we develop the prototypical verbalizer for mapping the predicted visual token with implicit semantics to explicit downstream labels. To our best knowledge, VPTM is the first visual prompt method on the generative pre-trained visual model, which achieves consistency between pre-training and downstream visual classification by task reformulation. Experiments show that VPTM outperforms other visual prompt methods and achieves excellent efficiency. Moreover, the task consistency of VPTM contributes to the robustness against prompt location, prompt length and prototype dimension, and could be deployed uniformly.
}

\keywords{Prompt learning, generative pre-trained model, task reformulation, prototypical verbalizer.}

\maketitle

\section{Introduction}
\label{sec:intro}
\begin{figure*}[tb!]
  \centering
  \includegraphics[width=15cm]{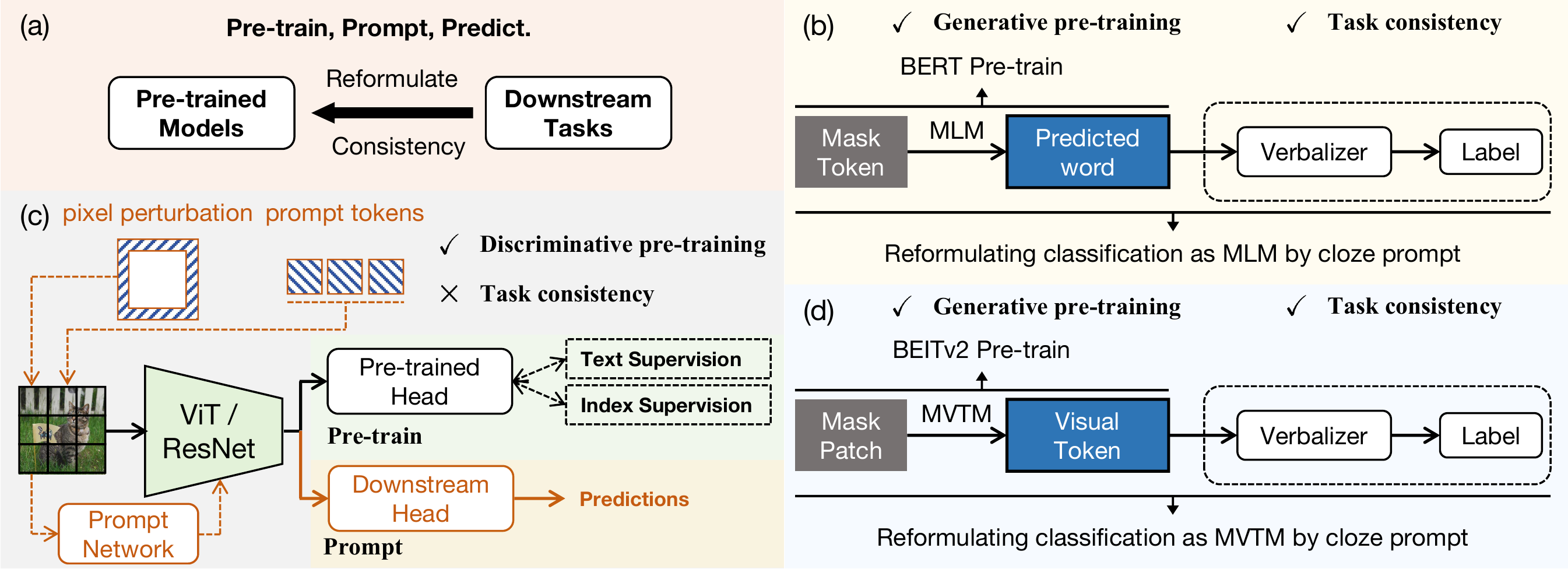}
  \caption{\textbf{Top.} Prompt learning in NLP reformulates downstream tasks as \emph{generative} pre-training tasks to keep consistency. Taking BERT~\citep{lewis2020bart} as an example, the downstream classification task is reformulated as the pre-trained masked language modeling (MLM) task by cloze prompt. The predicted word in the masked place is then mapped to downstream labels by the verbalizer. \textbf{Mid.} Current visual prompt learning methods are almost designed on discriminative pre-trained models, including supervised pre-trained ViT~\citep{dosovitskiy2020image} and ResNets~\citep{he2016deep} with index supervision, contrastively pre-trained ViT in CLIP~\citep{radford2021learning} with text supervision. There are three typical ways in current visual prompt learning methods: i) concatenating prompt tokens to the image patch tokens (VPT~\citep{jia2022vpt}); ii) adding pixel-wise perturbation as prompts (VP~\citep{bahng2022visual}, ILM-VP~\citep{chen2022understanding}, EVP~\citep{wu2022unleashing}); iii) learning prompt network (PGN~\citep{loedeman2022prompt}). When prompt tuning in downstream applications, the objectives of these methods are different from the pre-training ones, causing the lack of task consistency. \textbf{Bottom.} To design a visual prompt learning method on generative pre-trained model as well as keeping the task consistency, our method reformulates the downstream visual classification task as the generative masked visual token modeling (MVTM) pre-training task in BEITv2~\citep{peng2022beit}. The predicted visual token in masked place is mapped to downstream labels by the verbalizer.}
  \label{fig:compa_overview}
\end{figure*}
Large-scale pre-trained models (PMs) have greatly promoted the development in the computer vision (CV) field~\citep{he2020momentum, caron2021emerging, chen2021empirical, grill2020bootstrap}. The common paradigm is firstly pre-training, then fine-tuning the entire model with different task-specific objectives in downstream applications, which is prohibitive. Such a significant problem also arises in the natural language processing (NLP) field and is even trickier due to the larger scales of pre-trained language models. 

To mitigate the issue in the paradigm, namely ``pre-train, then fine-tune"~\citep{radford2018improving, dong2019unified, yang2019xlnet, lewis2020bart} in NLP, a new paradigm, namely ``pre-train, prompt, then predict"~\citep{liu2021pre} has been proposed~\citep{petroni2019language, raffel2020exploring, gao2021making}. Based on the generative pre-trained language models, \emph{the core technology is to reformulate downstream tasks to be the same form as the pre-training language modeling tasks}, as shown on the top of Fig.~\ref{fig:compa_overview}. In this way, when PMs are applied to downstream tasks, the knowledge of PMs can be naturally exploited with same objectives as in pre-training tasks, and contributes to better performance stably.

\begin{figure*}[tb!]
  \centering
  \includegraphics[width=16cm]{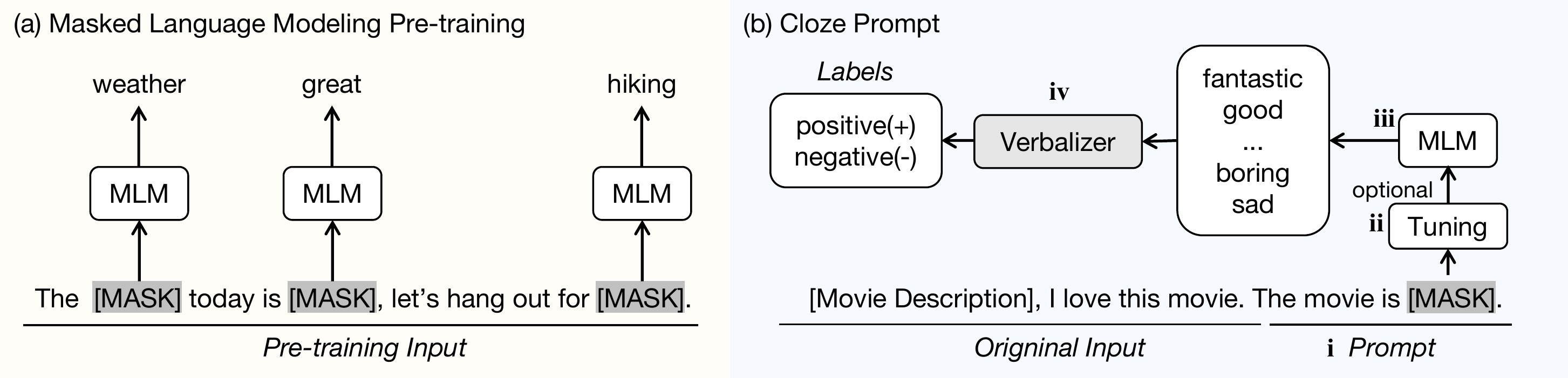}
  \caption{(a) Language models such as BERT~\citep{kenton2019bert} take the masked language modeling (MLM) task for pre-training. (b) To be consistent with the MLM pre-training task, the text classification task is reformulated by the cloze prompt. The verbalizer maps the words predicted from the mask to downstream labels.}
  \label{fig:NLP_prompt}
\end{figure*}

Taking the masked language modeling (MLM) pre-training task as an example~\citep{kenton2019bert, zhang2019ernie}, as shown in Fig.~\ref{fig:NLP_prompt} (a), classification tasks in NLP are usually reformulated by cloze prompt, which follows four steps~\citep{kumar2016ask, mccann2018natural} in Fig.~\ref{fig:NLP_prompt} (b): (i) adding prompts with masks to the original input; (ii) performing prompt tuning (optional); (iii) predicting the word in the masked place from the vocabulary by MLM; (iv) mapping the predicted word to downstream labels using verbalizer. Specifically, the predicted words are usually not the downstream labels, thus verbalizer~\citep{hu2022knowledgeable, schick2020automatically, holtzman2021surface, gao2021making} is devised to establish connections between them, e.g., the verbalizer in Fig.~\ref{fig:NLP_prompt} (b) maps ``fantastic, good" to ``positive". As such, prompt learning can be well applied in solving tasks such as text classification, named entity recognition, etc.

Witnessing the success of prompt learning in NLP, researchers introduce prompt learning into vision applications~\citep{nie2022pro, loedeman2022prompt, sohn2022visual}. VPT~\citep{jia2022vpt} prepends a few parameters as prompts to the input sequence of ViT~\citep{dosovitskiy2020image}, which has been supervised pre-trained on ImageNet-21k~\citep{deng2009imagenet} in a discriminative way. Visual prompting (VP)~\citep{bahng2022visual} modifies the pixel space with learnable parameters to perform visual prompt learning on CLIP~\citep{radford2021learning}, which has been pre-trained by contrastive learning. Till now, the current visual prompt learning methods~\citep{loedeman2022prompt, sohn2022visual, wu2022unleashing, nie2022pro} are all designed on discriminative pre-trained models shown in the middle of Fig.~\ref{fig:compa_overview}. \emph{There lacks prompt learning method carefully designed for the generative pre-trained visual model.} Particularly, regardless of the efforts paid on adding prompts in the input space~\citep{jia2022vpt, bahng2022visual, wu2022unleashing, chen2022understanding}, learning prompt networks~\citep{loedeman2022prompt} or designing prompt blocks~\citep{nie2022pro}, \emph{unifying the forms of pre-training and downstream applications by task reformulation to achieve consistency remains unexplored.} In view of the improved performance, efficiency and stability brought by the task consistency of prompt learning in NLP, we aim at the task consistency of generative visual prompt learning by inheriting the generative pre-training task to the downstream visual classification task.

For this, based on the generative model BEITv2~\citep{peng2022beit}, in which the visual tokens of masked patches are predicted from the codebook in pre-training inspired by BERT~\citep{kenton2019bert}, we propose Visual Prompt learning as masked visual Token Modeling (VPTM) for the visual classification task, as shown in the bottom of Fig.~\ref{fig:compa_overview}. Specifically, we concatenate continuous prompts and pre-trained mask token to the input sequence in prompt tuning. \emph{The downstream classification is achieved by mapping the prediction in the masked place, as in the pre-training phase, to the downstream labels by the verbalizer}. Considering that the semantics of visual tokens are implicit and constructing a verbalizer manually is intractable, we introduce the prototypical verbalizer into VPTM inspired by NLP~\citep{wei2022eliciting, cui2022prototypical}.

Experimentally, VPTM outperforms other visual prompt learning methods~\citep{jia2022vpt, bahng2022visual, chen2022understanding, loedeman2022prompt, wu2022unleashing} with better efficiency. Extensive experiments show the consistency between pre-training and downstream visual classification contributes to the robustness against learning strategies for different datasets, prompt locations, prompt length, and prototype dimensions. As a result, the VPTM equips with the capability of unified development. Our contributions include:

1) We propose a visual prompt learning method, which reformulates the visual classification as the generative masked visual token modeling task. To the best of our knowledge, it is the first visual prompt learning method on the generative pre-trained visual model and achieves a close match between the forms of downstream and pre-training tasks.

2) For mapping from predicted visual tokens with implicit semantics to downstream labels, we introduce the prototypical verbalizer into the vision area to construct the mapping rule, instead of manual construction.

\section{Related Work}
\subsection{Prompt Learning in NLP}
\textbf{Reformulating downstream tasks as pre-training tasks.} Considering a limited application of fine-tuning the entire large-scale pre-trained model~\citep{radford2018improving, brown2020language, kenton2019bert} for downstream tasks, and the impaired performance due to the gap between pre-trained and fine-tuning tasks, researchers in NLP initially proposed prompt learning~\citep{petroni2019language, raffel2020exploring, gao2021making}, in which the downstream tasks are reformulated in the same form as the pre-training ones, and only a few additional prompt-relevant parameters are optimized~\citep{petroni2019language, raffel2020exploring, gao2021making}. Specifically, based on the generative \textit{masked language modeling} pre-training task~\citep{kenton2019bert}, the text classification~\citep{lester2021power, schick2021exploiting, schick2020automatically}, named entity recognition~\citep{cui2021template} and commonsense reasoning~\citep{ettinger2020bert} are transformed into \textit{cloze prompt}, in which prompts with the mask token are added to the original input. The tokens predicted in masked places by the pre-training task are then mapped to the answers by the verbalizer. Based on the generative \textit{casual language modeling} pre-training task~\citep{brown2020language}, the question answering~\citep{khashabi2020unifiedqa,jiang2020can}, text generation~\citep{brown2020language,schick2020few,li2021prefix} and automatic evaluation of text generation~\citep{yuan2021bartscore} are reformulated as \textit{prefix prompt}, in which a prefix string is prepended to the original input for autoregressive answer text generation as in pre-training. \emph{Our method is inspired by the core idea in prompt learning, i.e., reformulating the downstream tasks as the pre-training task to keep their consistency.}

\textbf{Verbalizer. } In the cloze prompt, the predicted words in masked places are usually not the actual labels. To map the predicted words to labels, handcrafted verbalizer~\citep{schick2020automatically, schick2021exploiting} was initially proposed by manually designed rules. To avoid the expert dependence and prediction bias in handcrafted verbalizer, the method~\citep{liu2021gpt} uses gradient descent to search the mapping. KPT~\citep{hu2022knowledgeable} incorporates external knowledge bases into verbalizer for text classification. Soft verbalizer~\citep{hambardzumyan2021warp, zhang2021differentiable} regards each label as a trainable token, which is optimized together with prompt tuning. Prototypical verbalizer~\citep{wei2022eliciting, cui2022prototypical} learns prototype vectors, which represent classes, as the verbalizer. The similarity between masked embedding and prototypes is adopted as the classification rule, i.e., the mapping from generated words to downstream labels. However, the semantic meaning of visual token in codebooks is implicit. Manually designing the mapping rule requires the explicit semantic meaning as the language words equipped with. Thus, it is inapplicable in visual prompt learning. Inspired by the prototypical verbalizer, we introduce it into our method to solve the problem of constructing mapping between visual tokens with implicit meaning and downstream labels.

\subsection{Visual Prompt Learning}
As an effective and efficient alternative technology of fine-tuning, prompt learning has been introduced into unimodal vision area~\citep{li2022learning, gao2022visual, zheng2022prompt, loedeman2022prompt, sohn2022visual, yang2023visual, liu2022prompt, huang2023diversity, zhang2022neural, wang2022dualprompt}. There are three typical ways in visual prompt learning. The first is to concatenate learnable prompt tokens to the image patch sequences in Transformer-based visual models~\citep{dosovitskiy2020image}. For example, VPT~\citep{jia2022vpt} is a representative visual prompt method based on supervised pre-trained ViT, which optimizes the prepended prompt-relevant parameters together with the newly added classification head for downstream visual tasks including classification. LPT~\citep{dong2022lpt} concatenates shared prompts for all classes and group-specific prompts to image patch sequence for long-tailed image classification. The second typical way is to learn pixel perturbation as prompts. Visual prompting (VP)~\citep{bahng2022visual} introduces learnable pixel perturbation on the image encoder of CLIP~\citep{radford2021learning} as prompts to be optimized. Based on VP~\citep{bahng2022visual}, EVP~\citep{wu2022unleashing} learns more diversified pixel perturbation as prompts by applying data augmentations. The third typical way is to learn prompt networks. Pro-tuning~\citep{nie2022pro} adapts pre-trained ResNets~\citep{he2016deep} to downstream tasks by introducing lightweight prompt blocks. PGN~\citep{loedeman2022prompt} learns a network to generate the prompts conditioned on input image. However, these methods are all designed on the discriminative pre-trained visual models. Prompt learning on generative pre-trained visual models by keeping consistency between pre-training and downstream applications remains unexplored. In this paper, we concentrate on \emph{prompt learning on generative pre-trained unimodal vision model, and reformulating the downstream visual classification task as the pre-training one to achieve task consistency.}

\subsection{Masked Modeling Pre-training}
Masked language modeling (MLM) is a representative generative pre-training task in language models such as BERT~\citep{kenton2019bert} and ERNIE~\citep{zhang2019ernie}. With masking pieces of the inputted sentences, it aims at predicting the masked text pieces based on the context, as a result of which the comprehensive understanding ability is equipped for the pre-trained model. Motivated by MLM, masked image modeling (MIM) was proposed to boost the visual pre-trained models~\citep{he2022masked, zhou2021image, chen2022context}. BEIT~\citep{bao2021beit} was proposed to predict the visual tokens, which are tokenized by the codebook of DALL-E~\citep{ramesh2021zero}, of the masked patches. However, BEIT~\citep{bao2021beit} is limited to learn the low-level features in the codebook of DALL-E. To solve the drawback of semantic-less representations of BEIT~\citep{bao2021beit}, BEITv2~\citep{peng2022beit} was then proposed with a semantic-rich tokenizer guided by CLIP~\citep{radford2021learning} or DINO~\citep{caron2021emerging}. The pre-training strategies of BEIT and BEITv2 are the same. As the visual tokens are supposed to be equipped with high-level semantics as the language words in the cloze prompt, we propose the visual prompt learning in consistency with masked visual token modeling on BEITv2.

\section{Preliminary}
Before elaborating our method, we first introduce the masked language modeling (MLM) pre-training task in Section~\ref{sec:pre_mlm}. The cloze prompt in NLP is then presented in Section~\ref{sec:pre_cloze}, which inspires us significantly, including motivation and technical design. After that, we introduce the masked visual token modeling (MVTM) pre-training task of BEIT-series models~\citep{bao2021beit, peng2022beit} in Section~\ref{sec:pre_mvtm}, which is the foundation of our method.

\subsection{Masked Language Modeling Pre-training}
\label{sec:pre_mlm}
Masked language modeling (MLM) is a representative pre-training task in language models, e.g., BERT~\citep{kenton2019bert}. Given a tokenized sentence $\boldsymbol{s}=[t_1, t_2, ..., t_n]$ consisting of $n$ tokens $t_i$ from a corpora $\mathcal{C}$, MLM randomly masks $15\%$ tokens from $\boldsymbol{s}$, and gets the masked sentence $\boldsymbol{s}^\mathcal{M}$, in which $\mathcal{M}$ is the index set of masked tokens. Based on $\boldsymbol{s}^\mathcal{M}$, MLM predicts the original tokens $v$ of the $\tt [MASK]$ token, as shown in Fig.~\ref{fig:NLP_prompt} (a). The objective of the MLM pre-training task is formulated as:
\begin{equation}
\mathcal{L} = - \sum\limits_{\boldsymbol{s} \in \mathcal{C}}\sum\limits_{i \in \mathcal{M}}  \log{p(v_i|\boldsymbol{s}^\mathcal{M})}.
\label{eq:mlm}
\end{equation}

After the MLM pre-training, the model is equipped with a comprehensive language understanding based on the context, which lays the foundation for the cloze prompt, a classical prompt format in NLP.

\subsection{Cloze prompt in NLP}
\label{sec:pre_cloze}
In the ``pre-train, then fine-tune" paradigm, the pre-trained language models (LMs) are applied to downstream tasks by fine-tuning all parameters according to different task-specific objectives, which causes expensive computation costs and task discrepancy. Taking the MLM pre-trained model BERT~\citep{kenton2019bert} as an example, if the downstream tasks can be resolved in the cloze format, in which the desired output is predicted from the masked places given the input as the context, the LMs can be easily applied to different downstream tasks in the uniform way they have been pre-trained.

Based on this idea, the cloze prompt is designed on MLM pre-trained models~\citep{kenton2019bert, zhang2019ernie} to fit downstream tasks that require \textit{word-level} outputs, as shown in Fig.~\ref{fig:NLP_prompt} (b). The key design is to transform the task-specific predictions as the generative predictions in the $\tt[MASK]$ token. Given the original input sentence tokenized as $\boldsymbol{s}=[t_1, t_2, ..., t_n]$, the pre-trained $\tt[CLS]$ token, $\tt[MASK]$ token, and prompt text $\boldsymbol{P_T}$ are concatenated with $\boldsymbol{s}$ to get the transformed input $\boldsymbol{s}^t$:
\begin{equation}
\boldsymbol{s}^t = [\tt[CLS], t_1, t_2, ..., t_n, \boldsymbol{P_T}, \tt{[MASK]}].
\label{eq:cloze}
\end{equation}

The prompt text $\boldsymbol{P_T}$ can be categorized into two types: unoptimizable discrete prompts that consist of real words in the vocabulary, and optimizable continuous prompts from virtual parameterized embeddings. The unoptimizable discrete prompts are usually adopted at zero-shot predictions, while the optimizable continuous prompts are usually tuned with the pre-trained model kept frozen specifically on downstream datasets, known as ``prompt tuning".

After getting the predicted word $Y_p$ from the $\tt[MASK]$ token in $\boldsymbol{s}^t$ by MLM, the downstream task is finally achieved through the mapping from $Y_p$ to the final labels $Y$ using the verbalizer $\boldsymbol{V}$, which is formulated as:
\begin{equation}
Y = \boldsymbol{V}(Y_p).
\label{eq:vb_nlp}
\end{equation}
\begin{figure*}[tb!]
  \centering
  \includegraphics[width=16cm]{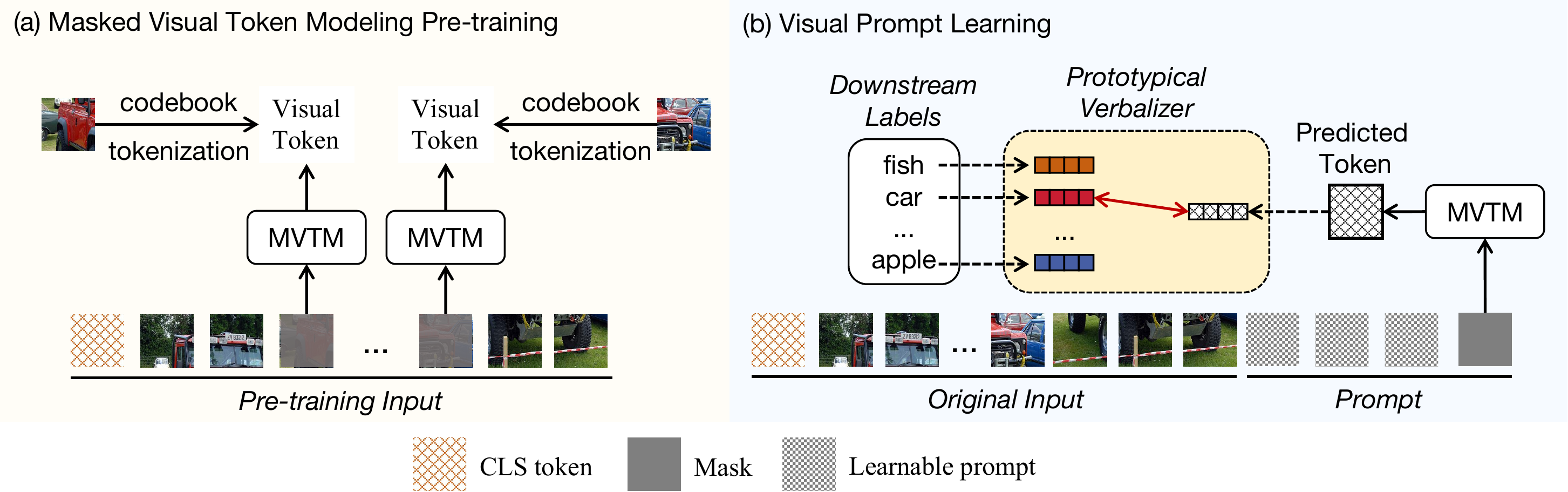}
  \caption{(a) BEIT-series models~\citep{bao2021beit, peng2022beit} are pre-trained by predicting the visual tokens of the masked patches. (b) The proposed VPTM reformulates the visual classification task as the generative masked visual token modeling (MVTM) task with pre-trained BEITv2. The proposed prototypical verbalizer constructs the connection between predicted visual token with implicit semantics and the downstream labels. The positions that the prompts and masks relative to the original input is ablating studied in experiments.}
  \label{fig:our_method}
\end{figure*}

\subsection{Masked Visual Token Modeling Pre-training}
\label{sec:pre_mvtm}
Following BERT~\citep{kenton2019bert}, BEIT-series (\textit{BERT Pre-Training of Image Transformers}) models~\citep{bao2021beit, peng2022beit} are pre-trained by the generative masked visual token modeling (MVTM). Specifically, each image $\boldsymbol{x}$ in dataset $\mathcal{D}$ is processed into patches. Then, all patches are tokenized into visual tokens within the codebook. In the pre-training phase, part of the patches indexed within a set $\mathcal{M}$ is replaced with {\tt [MASK]}. The model is trained to predict the visual token $z$ of the {\tt [MASK]} patches in the masked image $\boldsymbol{x}^\mathcal{M}$, as shown in Fig.~\ref{fig:our_method} (a). The pre-training loss is:
\begin{equation}
\mathcal{L} = - \sum\limits_{\boldsymbol{x} \in \mathcal{D}}\sum\limits_{i \in \mathcal{M}}  \log{p(z_i|\boldsymbol{x}^\mathcal{M})}.
\label{eq:mvtm}
\end{equation}

Based on the high similarity between MLM and MVTM pre-training tasks (also can be seen from Eq.~\ref{eq:mlm} and Eq.~\ref{eq:mvtm}), and inspired by the cloze prompt in NLP, we propose the visual prompt learning method, namely VPTM, which is the first prompt method designed on the generative pre-trained visual model as well as keeping the task consistency.

\section{Method}
In this section, we elaborate the proposed method VPTM, which reformulates the downstream visual classification task as the generative masked visual token modeling (MVTM) pre-training task. The overview of our method is shown in Fig.~\ref{fig:our_method}. We first introduce the task reformulation in Section~\ref{sec:overall_pipeline}. The prototypical verbalizer for mapping the predicted visual token to downstream labels is devised in Section~\ref{sec:verbalizer}.

\subsection{Visual Prompt Learning as MVTM}
\label{sec:overall_pipeline}
Based on the masked visual token modeling (MVTM) pre-training in Fig.~\ref{fig:our_method} (a), we propose the visual prompt learning method VPTM, as shown in Fig.~\ref{fig:our_method} (b). Each image $\boldsymbol{x} \in \mathbb{R}^{H\times W \times C}$ is firstly processed into $N$ patches $\{\boldsymbol{x}_i^p\}_{i=1}^N$ with $N=HW/P^2$ and the patch size as $P\times P$. $H, W$ is the image resolution, $C$ is the number of channels. The patches are transformed into $d$-dimension patch embeddings $\boldsymbol{e}_i\in\mathbb{R}^{d}$ with positional encoding by the pre-trained BEITv2~\citep{peng2022beit} through an embedding function ${\tt Embed}_{pre}$:
\begin{equation}
\boldsymbol{e}_i = {\tt Embed}_{pre} (\boldsymbol{x}_i^p), i=1,2,...,N.
\label{eq:patch_embed}
\end{equation}

The classification token $\boldsymbol{e}_{{\tt [CLS]}}$ of the pre-trained model is then prepended in the front of the sequence of patch embeddings to obtain the original input, and keep the same token interactions as in pre-training.
 
\textit{Instead of adding a new prediction head on the classification token to perform downstream tasks} as current visual prompt methods~\citep{jia2022vpt, bahng2022visual, chen2022understanding} do, \textit{our method reformulates the visual classification task as the MVTM task}. To preserve the image information, we concatenate the {\tt [MASK]} token $\boldsymbol{e}_{\tt [MASK]}$, which is also from the pre-trained model, to the original input sequence. Besides, for continuously tuning towards downstream datasets, we insert additional $N_p$ parameterized learnable prompts $\boldsymbol{p}_{i}\in\mathbb{R}^{d}, i \in [1, N_p]$ into the sequence, the dimension of $\boldsymbol{p}_{i}$ is the same as the hidden dimension of the pre-trained BEITv2~\citep{peng2022beit}. Then, we obtain the final input sequence $\boldsymbol{H}_{vp}$:
\begin{equation}
\boldsymbol{H}_{vp} = [\boldsymbol{e}_{{\tt [CLS]}}, \boldsymbol{e}_1, ..., \boldsymbol{e}_N, \boldsymbol{p}_1, ..., \boldsymbol{p}_{N_p}, \boldsymbol{e}_{\tt [MASK]}].
\label{eq:input_vp}
\end{equation}

The positions of the prompts $\boldsymbol{p}_{i}$ and {\tt [MASK]} token $\boldsymbol{e}_{\tt [MASK]}$ related to the original image patches are ablating studied in Section~\ref{sec:ablate_pos} in experiments.

As such, the input sequence $\boldsymbol{H}_{vp}$ in Fig.~\ref{fig:our_method} (b) bottom is similar to the input sequence in the pre-training stage in Fig.~\ref{fig:our_method} (a), in which each sequence includes the {\tt [MASK]} token. The original input sequence in $\boldsymbol{H}_{vp}$ can be regarded as the context for predicting the visual token that supposed to be in the masked place by MVTM. Our method fulfills the cloze format prompt in vision area, which is similar to the cloze prompt in NLP, as shown in Fig.~\ref{fig:NLP_prompt} (b). After feeding $\boldsymbol{H}_{vp}$ into the pre-trained model, we get the embedding of the {\tt [MASK]} token denoted as $\boldsymbol{h}_{\tt [MASK]} \in \mathbb{R}^{d}$. To achieve visual classification by MVTM, the last step is to map the visual token to downstream labels, which is introduced in Section~\ref{sec:verbalizer}.
\subsection{Prototypical Verbalizer}
\label{sec:verbalizer}
In the vision area, the visual tokens are equipped with implicit semantic meaning. Designing the mapping rule from the visual token to downstream labels manually is intractable. To solve the mapping problem, we propose the prototypical verbalizer inspired by~\citep{wei2022eliciting, cui2022prototypical}.

For each class, we devise the corresponding learnable prototype vector $\boldsymbol{c}_k\in\mathbb{R}^{t} , k \in [1,N_C]$, in which $N_C$ is the number of downstream classes and $t$ is the dimension of the prototype vector. After getting the embedding of the {\tt [MASK]} token $\boldsymbol{h}_{\tt [MASK]} \in \mathbb{R}^{d}$, we project it into the prototype space using a linear function  $\mathcal{F}: \mathbb{R}^{d} \to \mathbb{R}^{t}$, then get vector $\boldsymbol{u_{\tt [MASK]}} \in \mathbb{R}^{t}$:
\begin{equation}
\boldsymbol{u_{\tt [MASK]}} = \mathcal{F}(\boldsymbol{h}_{\tt [MASK]}).
\label{eq:proj2proto}
\end{equation}

The mapping from the visual token to downstream labels is transformed as the similarity between vector $\boldsymbol{u_{\tt [MASK]}}$ and each prototype vector. The similarity between an image $\boldsymbol{x}_i$ and its class $C_i$ with prototype  $\boldsymbol{c}_i$ is thus calculated as:
\begin{equation}
{\tt sim}(\boldsymbol{x}_i,C_i)= \boldsymbol{u^i_{\tt [MASK]}} \cdot \boldsymbol{c}_i^T,
\label{eq:sim}
\end{equation}
where $T$ is the transpose manipulation. 

For a batch of $N$ images, the loss is:
\begin{equation}
\begin{aligned}
\mathcal{L}_{vp} &= -\frac{1}{N}\sum_{i=1}^{N}{\rm log}\frac{{\rm exp}({\tt sim}(\boldsymbol{x}_i,C_i))}{ { \sum_{k=1}^{N_C}{\rm exp}({\tt sim}(\boldsymbol{x}_i,C_k))} }\\
&= -\frac{1}{N}\sum_{i=1}^{N}{\rm log}\frac{{\rm exp}(\boldsymbol{u^i_{\tt [MASK]}} \cdot \boldsymbol{c}_i^T)}{ { \sum_{k=1}^{N_C}{\rm exp}(\boldsymbol{u^i_{\tt [MASK]}} \cdot \boldsymbol{c}_k^T)} }.
\label{eq:loss_vp}
\end{aligned}
\end{equation}

Overall, by inheriting the MVTM task, the learnable visual prompts and class prototypes are optimized to obtain the dataset-specific visual prompts and construct the mapping relationship from the prediction in the {\tt [MASK]} place to downstream labels. In the prompt tuning phase, all parameters in the pre-trained vision model are kept frozen.

\section{Experiments}
\subsection{Datasets}
To evaluate the performance of the proposed VPTM, we select 10 datasets in our experiments. The visual prompts are learned on the training sets and evaluated on the test sets. In the datasets, CIFAR100~\citep{krizhevsky2009learning} includes 100 classes, CIFAR10~\citep{krizhevsky2009learning} includes 10 classes, Oxford Flowers102~\citep{nilsback2008automated} includes 102 classes, Food101~\citep{bossard2014food} includes 101 classes, EuroSAT~\citep{helber2019eurosat} includes 10 classes, SVHN~\citep{netzer2011reading} includes 10 classes, Oxford Pets~\citep{parkhi2012cats} includes 37 classes, DTD~\citep{cimpoi2014describing} includes 47 classes, Resisc45~\citep{cheng2017remote} includes 45 classes, Patch Camelyon (PatchCame)~\citep{veeling2018rotation} includes 2 classes.
\subsection{Baseline Methods}
We compare our method with other downstream fine-tuning and visual prompt methods as baselines:

(1) \emph{Fine tune (FT).} Optimizing the entire model.

(2) \emph{Linear probe (FP).} Only optimizing the classification head on the {\tt [CLS]} token.

(3) \emph{Prompting on the image encoder of CLIP.} As the codebook of BEITv2 is distilled from CLIP, here we set some prompting methods designed on the image encoder of CLIP as baselines for direct comparisons. They are: a) fine-tuning CLIP; b) linear probing on CLIP; c) using textual prompt (TP); d) TP + visual prompt (VP)~\citep{bahng2022visual}, which adds perturbation on the pixel; e) TP + PGN~\citep{loedeman2022prompt}, which generates prompts for input; f) EVP~\citep{wu2022unleashing}, which adds prompts on the pixel with improved generalization; g) ILM-VP~\citep{chen2022understanding}, which adds prompts on the pixel and learns a label mapping.

(4) \emph{Visual prompt tuning (VPT)}~\citep{jia2022vpt}. Optimizing parameters in prompts and the newly added classification head for the downstream task. The prompts are prepended only at the first layer of the model. The classification head is devised on the {\tt [CLS]} token of the last layer of the model.

Additionally, our method focuses on \emph{unimodal visual prompt learning}, we could not compare with multi-modal prompt methods performed on texts, such as CoOp~\citep{zhou2022learning} and CoCoOp~\citep{zhou2022conditional}. We also could not compare with multi-modal prompt methods that the visual and textual prompts are learned jointly and could not be separated, including UPT~\citep{zang2022unified}, MaPLe~\citep{khattak2022maple}.
\begin{table*}[t!]
  \centering
  \caption{The accuracy comparisons between our method and baseline methods. LP: Linear probe. FT: Fine tune. TP: Textual prompt.}
  \footnotesize
  \label{tab:compa_with_baseline}
  \begin{tabular}{l@{}| p{0.8cm}<{\centering} p{0.6cm}<{\centering} p{0.6cm}<{\centering} p{0.6cm}<{\centering} p{0.6cm}<{\centering} p{0.6cm}<{\centering} p{0.6cm}<{\centering} p{0.6cm}<{\centering} p{0.6cm}<{\centering} p{1.2cm}<{\centering} | p{1.0cm}<{\centering} }
    \toprule
    \midrule
    Methods & CIFAR100 & CIFAR10 & Flowers & Food101 & EuroSAT & SVHN & Pets & DTD & Resisc45 & PatchCame & Average\\
    \midrule
    FT+BEITv2    & 92.35 & 98.94 & 99.01 & 92.70 & 99.28 & 98.11 & 93.92 & 81.65 & 97.51 & 87.42 & 94.09\\
    LP+BEITv2 & 78.13 & 93.50 & 85.57 & 81.91 & 96.76 & 63.11 & 90.16 & 69.47 & 89.75 & 81.03 & 82.94 \\
    \midrule
    FT+CLIP    & 82.10 & 95.80 & 97.40 & 80.50 & 97.90 & 95.70 & 88.50 & 72.30 & 94.40 &  N.R. & 89.40  \\
    LP+CLIP    & 80.00 & 95.00 & 96.90 & 84.60 & 95.30 & 65.40 & 89.20 & 74.60 & 66.00 &  N.R. & 83.00   \\
    \midrule
    TP+CLIP    & 63.10 & 89.00 & 61.90 & 79.80 & 40.00 &  5.10 & 85.90 & 43.00 & 42.40 &  N.R. & 56.69  \\
    TP+VP+CLIP & 75.30 & 84.20 & 70.30 & 78.90 & 96.40 & 88.40 & 85.00 & 57.10 & 81.40 &  N.R. & 79.67  \\
    TP+PGN+CLIP & 79.30 & 96.10 & \textbf{94.00} & 82.50 & 98.00 & 94.20 & \textbf{91.50} & 71.50  & 92.10 &  N.R.  & 88.80 \\ 
    EVP & \textbf{81.20} & 96.60 & 82.30 & 84.10 & \textbf{98.70} & 90.50 & 90.00 & 68.40 & \textbf{92.30} & N.R. & 87.12\\
    \textbf{VPTM (Ours, 100e)} & 80.15 & \textbf{98.20} & 91.35 & \textbf{84.28} & 98.57 & \textbf{91.76} & 91.25 & \textbf{76.81} & 90.86 & N.R. & \textbf{89.25} \\
    \midrule
    ILM-VP & N.R. & 94.40 & 83.70 & 79.10 & 96.90 & 91.20 & N.R. & 63.90 & N.R. & N.R. & 84.87 \\
    \textbf{VPTM (Ours, 100e)} & N.R. & \textbf{98.20} & \textbf{91.35} & \textbf{84.28} & \textbf{98.57} & \textbf{91.76} & N.R. & \textbf{76.81} & N.R. & N.R. & \textbf{90.16} \\
    \midrule
    % VPT~\citep{jia2022vpt}+BEITv2 (50e)  & \textbf{82.80} & 96.78 & 85.38 & \textbf{88.23} & 97.74 & 91.08 & 87.90 & 73.99 & 90.84 & 80.92 & 87.57 \\
    VPT+BEITv2 (100e)\,\,\,\,\, & \textbf{83.12} & 96.71 & 89.46 & \textbf{88.29} & 94.06 & 90.46 & 88.88 & 74.31 & 90.81 & 83.04 & 87.91 \\
    \textbf{VPTM (Ours, 50e)}  & 79.43 & 97.12 & 90.67 & 82.65 & 98.37 & 91.35 & \textbf{91.41} & 75.96 & \textbf{90.86} & 81.29 & 87.91 \\
    \textbf{VPTM (Ours, 100e)} & 80.15 & \textbf{98.20} & \textbf{91.35} & 84.28 & \textbf{98.57} & \textbf{91.76} & 91.25 & \textbf{76.81} & \textbf{90.86} & \textbf{83.91} & \textbf{88.71} \\
    \bottomrule
  \end{tabular}
\end{table*}
\begin{figure*}[tb!]
\begin{minipage}[b]{3.5cm}
  \centering
  \centerline{\includegraphics[width=3.5cm]{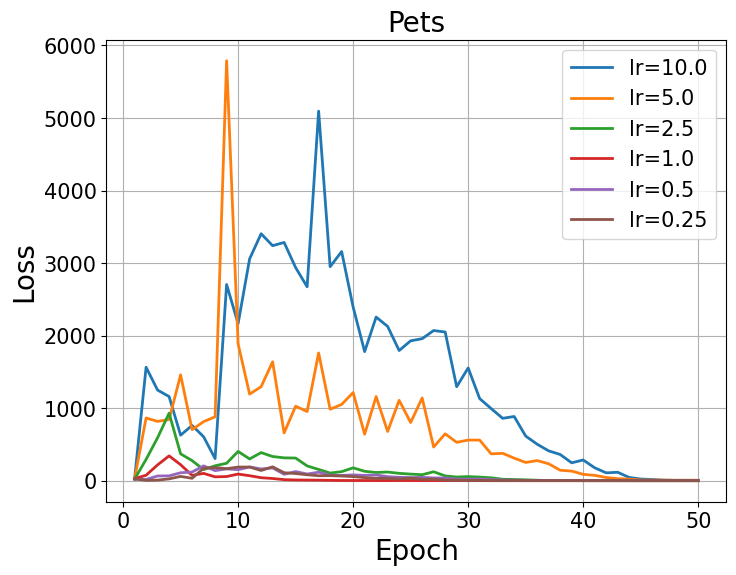}}
  \centerline{\footnotesize (a) Loss curve on Pets.}
\end{minipage}
\hfill
\begin{minipage}[b]{3.4cm}
  \centering
  \centerline{\includegraphics[width=3.4cm]{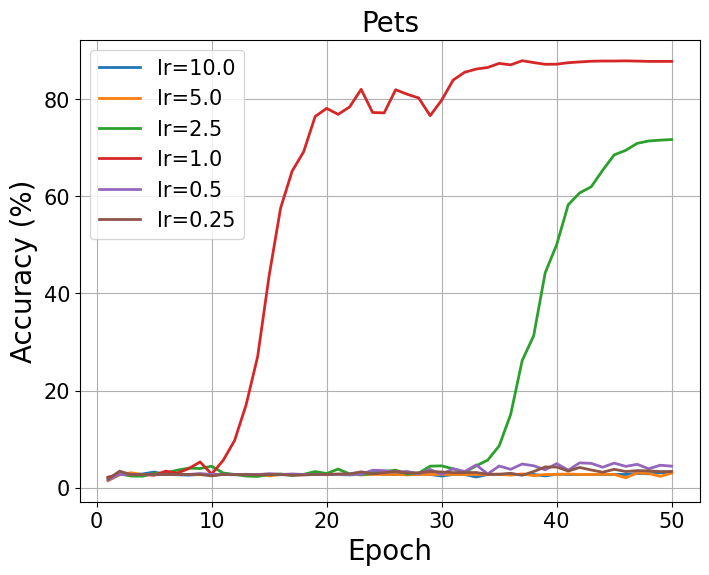}}
  \centerline{\footnotesize (b) Accuracy curve on Pets.}
\end{minipage}
\hfill
\begin{minipage}[b]{3.5cm}
  \centering
  \centerline{\includegraphics[width=3.5cm]{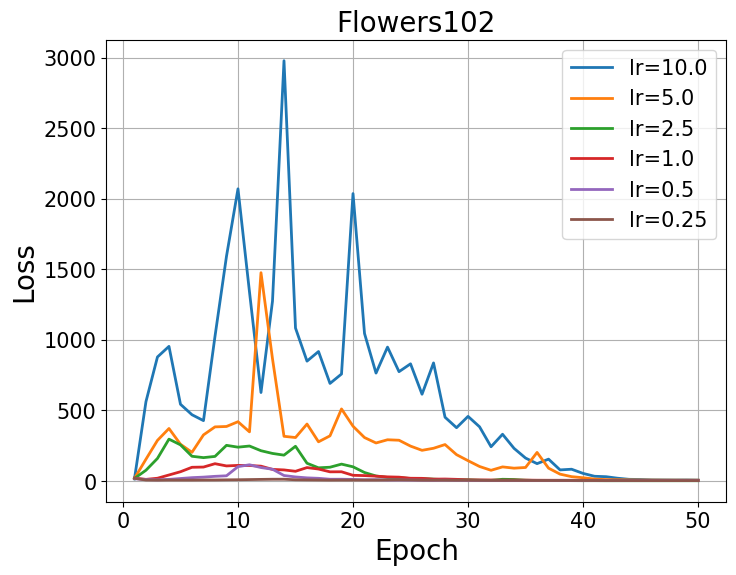}}
  \centerline{\footnotesize (c) Loss curve on Flowers. }
\end{minipage}
\hfill
\begin{minipage}[b]{3.4cm}
  \centering
  \centerline{\includegraphics[width=3.4cm]{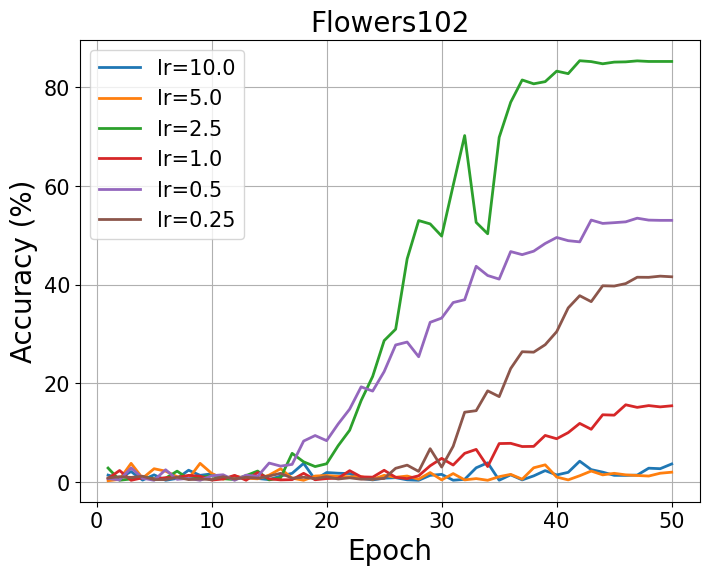}}
  \centerline{\footnotesize (d) Accuracy curve on Flowers. }
\end{minipage}
\hfill
\caption{Loss and accuracy curves with different hyperparameters in VPT~\citep{jia2022vpt}. The deployment of VPT is complex and time-consuming. The optimal learning rates on different datasets are different. One optimal learning rate for a dataset can cause failed learning on others.} 
\label{fig:complex_VPT}
\end{figure*}

\subsection{Implementation Details}
We experiment with the BEITv2~\citep{peng2022beit}, which has been pre-trained on ImageNet-1k~\citep{deng2009imagenet} by masked visual token modeling and the codebook is guided by CLIP~\citep{radford2021learning}. Main experiments of our method are performed for 50 epochs on NVIDIA Tesla V100 GPU with batch size 64. We use the AdamW optimizer with the weight decay set as 0.01 and the momentum set as 0.9. The base learning rate is 0.001. We use the cosine scheduler with 5 warm up epochs. The prompts and prototypes are all initialized as all zeros. Fine-tuning and linear probing on BEITv2~\citep{peng2022beit} are performed for 50 epochs following the official code~\footnote{https://github.com/microsoft/unilm/tree/master/beit2.}. We implement VPT on the weights of BEITv2 for a fair comparison~\footnote{VPT is officially performed on supervised pre-trained ViT~\citep{dosovitskiy2020image}.}. 
\subsection{Comparison to Baseline Methods}
The accuracy comparisons between our method and baselines are shown in Table~\ref{tab:compa_with_baseline}.

\textbf{Comparison with  FT \& LP on BEITv2.} By only optimizing $0.17\%$ parameters of the entire model on average, which will be discussed in Section~\ref{sec:param_effi}, our method is inferior to fine-tuning the entire model. On the other hand, our method consistently surpasses linear probe on all datasets more than about $5\%$ on average. Particularly, our method outperforms linear probe by nearly $30\%$ on SVHN dataset. These indicate that reformulating the classification task as theMVTM pre-training task in our method is superior to conventionally performing classification on the {\tt [CLS]} token in linear probe.

\textbf{Comparison with prompting methods on the image encoder of CLIP.} Our method achieves competitive performance compared with fine-tuning the entire CLIP~\citep{radford2021learning}. Concerning the results of VP~\citep{bahng2022visual} and PGN~\citep{loedeman2022prompt} combined with textual prompts, our method achieves the best average performance even without the assistance of texts. In addition, compared with EVP~\citep{wu2022unleashing} and ILM-VP~\citep{chen2022understanding}, which directly perform prompt learning on the image encoder of CLIP, our method still exhibits a great performance advantage. Hence, compared with the above baseline methods which add image perturbation in the pixel or learn a network to generate prompts, our method is more effective.

\textbf{Comparison with VPT~\citep{jia2022vpt}.} VPT and our method both concatenate learnable tokens with the original input patch sequence in prompt learning. Based on BEITv2, the only difference between VPT and our method is that our method inherits the pre-trained MVTM task by task reformulation, while VPT dose not. For a fair comparison, we implement VPT-shallow on BEITv2 for 100 epochs following the official setting of VPT~\citep{jia2022vpt}. The prompt length of each dataset is the same as the optimal setting in VPT. By tuning VPTM for only 50 epochs, our method could achieve comparable average performance to that of VPT under 100 epochs. The accuracy of VPTM are better on 7 out of 10 datasets. By tuning VPTM for 100 epochs, it outperforms VPT on the average accuracy by nearly 1 point, and achieves better performance on 8 out of 10 datasets.
\begin{table*}[t!]
  \centering
  \caption{Result comparisons between the proposed VPTM and VPT. The prompt length is set as optimal for each dataset, or 10 uniformly. The VPT is implemented using the same learning strategy as VPTM. The accuracy gap is the difference between the results achieved under the two sets of prompt length.}
  \footnotesize
  \label{tab:compa_VPT_uniform}
  \begin{tabular}{l@{} | l@{}| p{0.8cm}<{\centering} p{0.8cm}<{\centering} p{0.8cm}<{\centering} p{0.8cm}<{\centering} p{0.8cm}<{\centering} }
    \toprule
    \midrule
     & Methods & CIFAR100 & CIFAR10 & Food101 & EuroSAT & Pets \\
    \midrule
    \multirow{3}{*}{Prompt-Len (Optimal)}\,\,\,\,\, & VPT+ BEITv2 (100e) \,\,\,\,\, & 28.76 & 43.71 & 59.92 & 72.76 & 4.69 \\
    & \textbf{VPTM (Ours, 50e)}  & 79.43 & 97.12 & 82.65 & 98.37 & \textbf{91.41} \\
    & \textbf{VPTM (Ours, 100e)} & \textbf{80.15} & \textbf{98.20} & \textbf{84.28} & \textbf{98.57} & 91.25 \\
    \midrule
    \multirow{2}{*}{Prompt-Len ($N_p=10$)} \,\,\,\,\, &  VPT+ BEITv2 (100e) & 67.06 & 78.05 & 81.53 & 78.91 & 59.55 \\
    &\textbf{VPTM (Ours, 50e)} & \textbf{72.09} & \textbf{96.54} & \textbf{81.76} & \textbf{97.06} & \textbf{88.42} \\
    \midrule
    \multirow{2}{*}{Accuracy Gap} \,\,\,\,\, &  VPT+ BEITv2 (100e) & 38.30 & 34.34 & 21.61 & 6.15 & 54.86 \\
    &\textbf{VPTM (Ours, 50e)} & 7.34 & 0.58 & 0.89 & 1.31 & 2.99 \\
    \bottomrule
  \end{tabular}
\end{table*}

Furthermore, it is worth mentioning that VPT severely relies on the hyperparameters of learning strategy. As reported in the paper of VPT~\citep{jia2022vpt}, different datasets adopt different parameters of the learning rate and the weight decay. Given that, when implementing VPT on BEITv2, we search the optimal hyperparameters within $[50.0, 25.0, 10.0, 5.0, 2.5, 1.0, 0.5, 0.25, 0.1]$ for the learning rate and $[0.0, 0.01, 0.001, 0.0001]$ for the weight decay. Within the 36 sets of hyperparameters, it is observed that one set of hyperparameters suitable for one dataset usually causes failed learning on other datasets. Taking Oxford Pets and Oxford Flowers102 datasets as examples, as shown in Fig.~\ref{fig:complex_VPT}, different learning rates result in various performance on the training loss and test accuracy. Moreover, the accuracy on Flowers with the same learning rate as Pets (i.e., $1.0$) is about $70\%$ lower than that with the optimal learning rate of $2.5$. In comparison, the proposed VPTM are uniformly tuned with one set of hyperparameters across all datasets, and is easy to be deployed. We infer that the insensitivity of VPTM to hyperparameters specific on different datasets is due to its inheritance of the pre-training task. Based on the consistency between the pre-training and reformulated downstream tasks, the knowledge of the pre-trained model could be stably exploited in downstream applications. \textit{In short, we can effectively and stably gain the performance advantage without complex process in searching the optimal training hyperparameters.}
\begin{table*}[t!]
  \centering
  \caption{The comparisons between the proposed VPTM and VPT~\citep{jia2022vpt} on prototype dimension $t$, prompt length $N_p$, GFLOPs, and the ratio of the amount of tuned parameters to the entire parameters in the optimal setting.}
  \footnotesize
  \label{tab:params}
  \begin{tabular}{ p{0.6cm}<{\centering} l@{}| p{0.8cm}<{\centering} p{0.6cm}<{\centering} p{0.6cm}<{\centering} p{0.6cm}<{\centering} p{0.6cm}<{\centering} p{0.6cm}<{\centering} p{0.6cm}<{\centering} p{0.6cm}<{\centering} p{0.6cm}<{\centering} p{1.2cm}<{\centering}| p{1.0cm}<{\centering} }
    \toprule
    \midrule
    \multicolumn{2}{c|}{Index} & CIFAR100 & CIFAR10 & Flowers & Food101 & EuroSAT & SVHN & Pets & DTD & Resisc45 & PatchCame & Average\\
    \midrule
    \multirow{3}{*}{VPT} &Prompt-Len ($N_p$) \,\,\,\, & 100 & 100 & 200 & 100 & 50 & 200 & 50 & 1 & 50 & 5 & 85.6 \\
    & GFLOPS & 26.99 & 26.99 & 36.78 & 26.99 & 22.24 & 36.78 & 22.24 & 17.67 & 22.24 & 18.04 & 25.70 \\
    & Tuned / Total($\%$) \,\,\,\, & 0.18 & 0.10 & 0.27 & 0.18 & 0.05 & 0.19 & 0.08 & 0.04 & 0.09 & 0.01 & \textbf{0.12} \\
    \midrule
    \multirow{4}{*}{VPTM} & Proto-Dim ($t$) \,\,\,\,    & 128 & 128 & 256 & 128 & 64 & 64 & 64 & 256 & 128 & 256 & -- \\
    &Prompt-Len ($N_p$) \,\,\,\, & 20 & 100 & 20 & 20 & 100 & 100 & 50 & 10 & 50 & 20 & \textbf{49.0} \\
    & GFLOPS & 19.53 & 27.09 & 19.53 & 19.53 & 27.09 & 27.09 & 22.34 & 18.60 & 22.34 & 19.53 & \textbf{22.27}\\
    & Tuned / Total($\%$) \,\,\,\, & 0.14 & 0.19 & 0.23 & 0.14 & 0.14 & 0.14 & 0.10 & 0.23 & 0.15 & 0.23 & 0.17 \\
    \bottomrule
  \end{tabular}
\end{table*}

In addition, we make comparisons between our method and VPT under the same learning strategies, i.e., setting the learning rate, weight decay, and momentum as 0.001, 0.01, 0.9 for all datasets. By setting the prompt length as optimal for each dataset, i.e., keeping the same setting corresponding to that in Table 1 in the main paper, or 10 uniformly, the results are shown in Table~\ref{tab:compa_VPT_uniform}. Our method achieves the best performance and exhibits decisive superiority under the two settings of prompt length. Moreover,  compared with the results of VPT in Table~\ref{tab:compa_with_baseline}, under the optimal setting of the prompt length, the accuracy on CIFAR100 drops from 83.12 to 28.76 significantly. VPT is shown to be quite sensitive to the learning strategies per dataset, and unsatisfying results could be caused when adopting a unified learning strategy for all datasets.

Moreover, VPTM achieves a far lower accuracy gap between setting the prompt length as the optimal one and 10, compared with that of VPT. The proposed VPTM is shown to be much more robust against the prompt length than VPT, which will be further ablating studied in Section~\ref{sec:ablate_promptLen}. 

In conclusion, based on the BEITv2 model, reformulating visual classification as the masked visual token modeling pre-training to achieve task consistency in visual prompt learning is rational. Compared with VPT, the proposed VPTM is shown to be more effective in performance, more stable in tuning, and more robust against the prompt length. 

\subsection{Parameter Efficiency Validation}
\label{sec:param_effi}
\begin{table*}[t!]
  \centering
  \caption{The comparisons between the MLP-1, MLP-2 and the prototypical verbalizer (PV).}
  \footnotesize
  \label{tab:verba_compa}
  \begin{tabular}{ p{0.6cm}<{\centering} l@{}| p{0.8cm}<{\centering} p{0.6cm}<{\centering} p{0.6cm}<{\centering} p{0.6cm}<{\centering} p{0.6cm}<{\centering} p{0.6cm}<{\centering} p{0.6cm}<{\centering} p{0.6cm}<{\centering} p{0.6cm}<{\centering} p{1.2cm}<{\centering} | p{1.0cm}<{\centering} }
    \toprule
    \midrule
    \multicolumn{2}{c|}{Index} & CIFAR100 & CIFAR10 & Flowers & Food101 & EuroSAT & SVHN & Pets & DTD & Resisc45 & PatchCame & Average\\
    \midrule
    \multirow{2}{*}{MLP-1} &Tuned / Total($\%$) \,\,\,\, & 0.10 & 0.09 & 0.10 & 0.10 & 0.09 & 0.09 & 0.07 & 0.05 & 0.08 & 0.02 & 0.08 \\
    & Accuracy ($\%$) \,\,\,\, & 67.07 & 97.03 & 71.85 & 67.32 & 96.52 & 90.14 & 79.18 & 73.24 & 77.11 & 77.87 & 79.73 \\
    \midrule
    \multirow{2}{*}{MLP-2} &Tuned / Total($\%$)) \,\,\,\,    & 0.14 & 0.19 & 0.17 & 0.14 & 0.19 & 0.19 & 0.15 & 0.18 & 0.15 & 0.15 & 0.17 \\
    &Accuracy ($\%$) \,\,\,\, & 74.42 & 97.01 & 81.92 & 75.04 & 97.09 & 91.16 & 90.19 & 76.81 & 90.52 & 79.39 & 85.36 \\
    \midrule
    \multirow{2}{*}{PV (Ours)} &Tuned / Total($\%$) \,\,\,\, & 0.14 & 0.19 & 0.23 & 0.14 & 0.14 & 0.14 & 0.10 & 0.23 & 0.15 & 0.23 & 0.17 \\
    &Accuracy ($\%$) \,\,\,\, & 79.43 & 97.12 & 90.67 & 82.65 & 98.37 & 91.35 & 91.41 & 75.96 & 90.86 & 81.29 & \textbf{87.91} \\
    \bottomrule
  \end{tabular}
\end{table*}

\begin{figure*}[tb!]
\begin{minipage}[b]{3cm}
  \centering
  \centerline{\includegraphics[width=3cm]{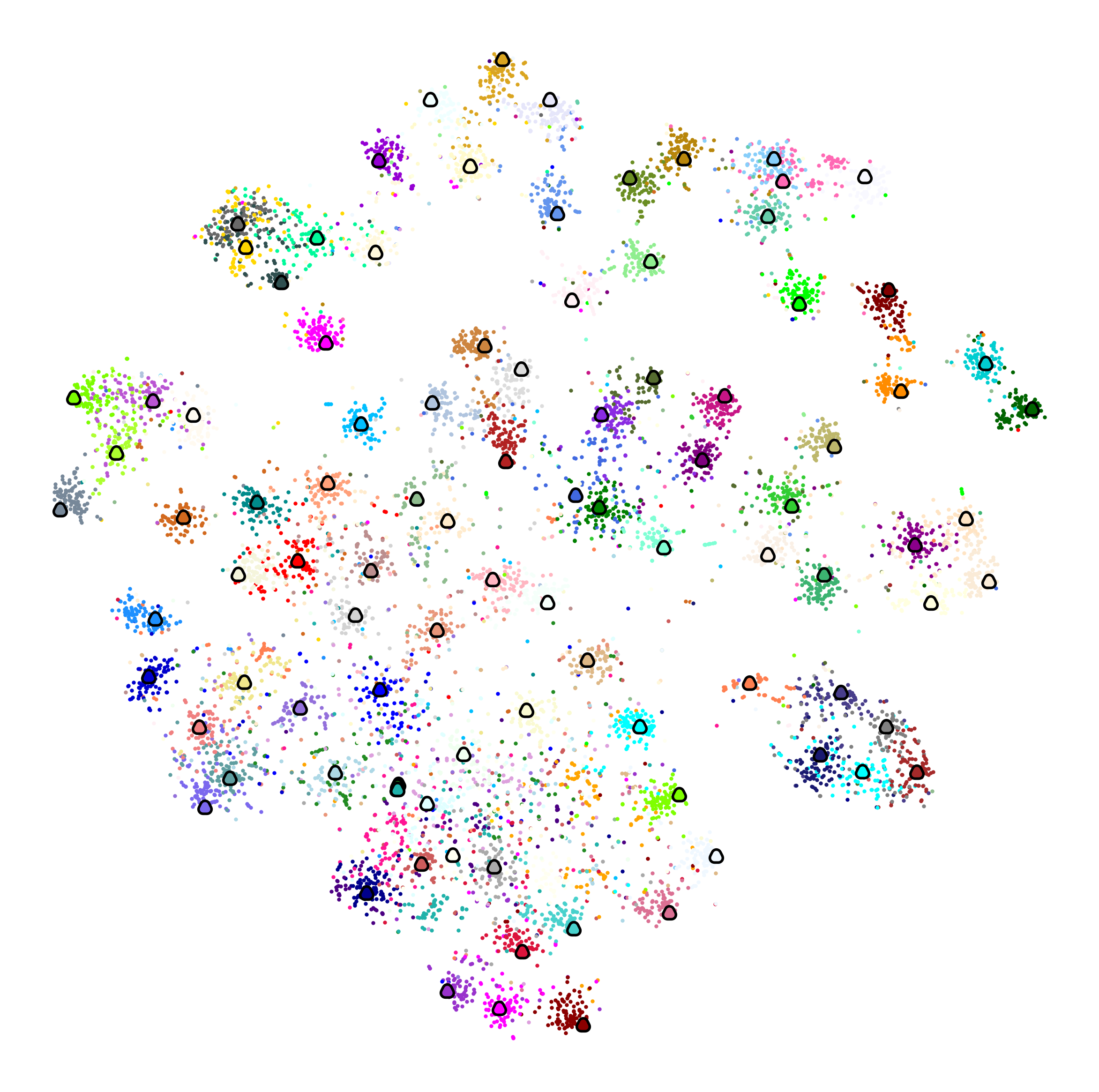}}
  \centerline{\footnotesize (a) CIFAR100.}
\end{minipage}
\hfill
\begin{minipage}[b]{3cm}
  \centering
  \centerline{\includegraphics[width=3cm]{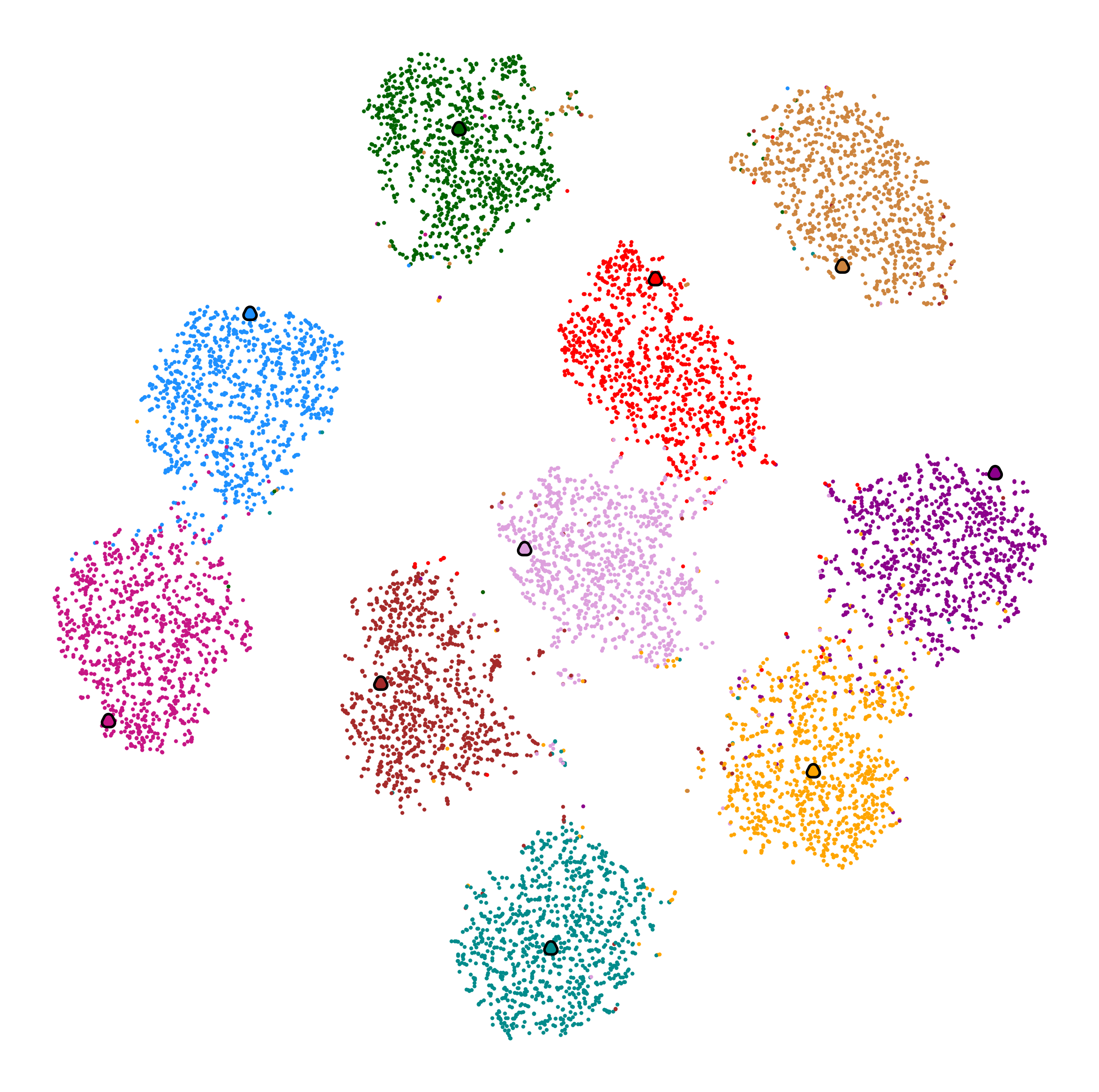}}
  \centerline{\footnotesize (b) CIFAR10.}
\end{minipage}
\hfill
\begin{minipage}[b]{3cm}
  \centering
  \centerline{\includegraphics[width=3cm]{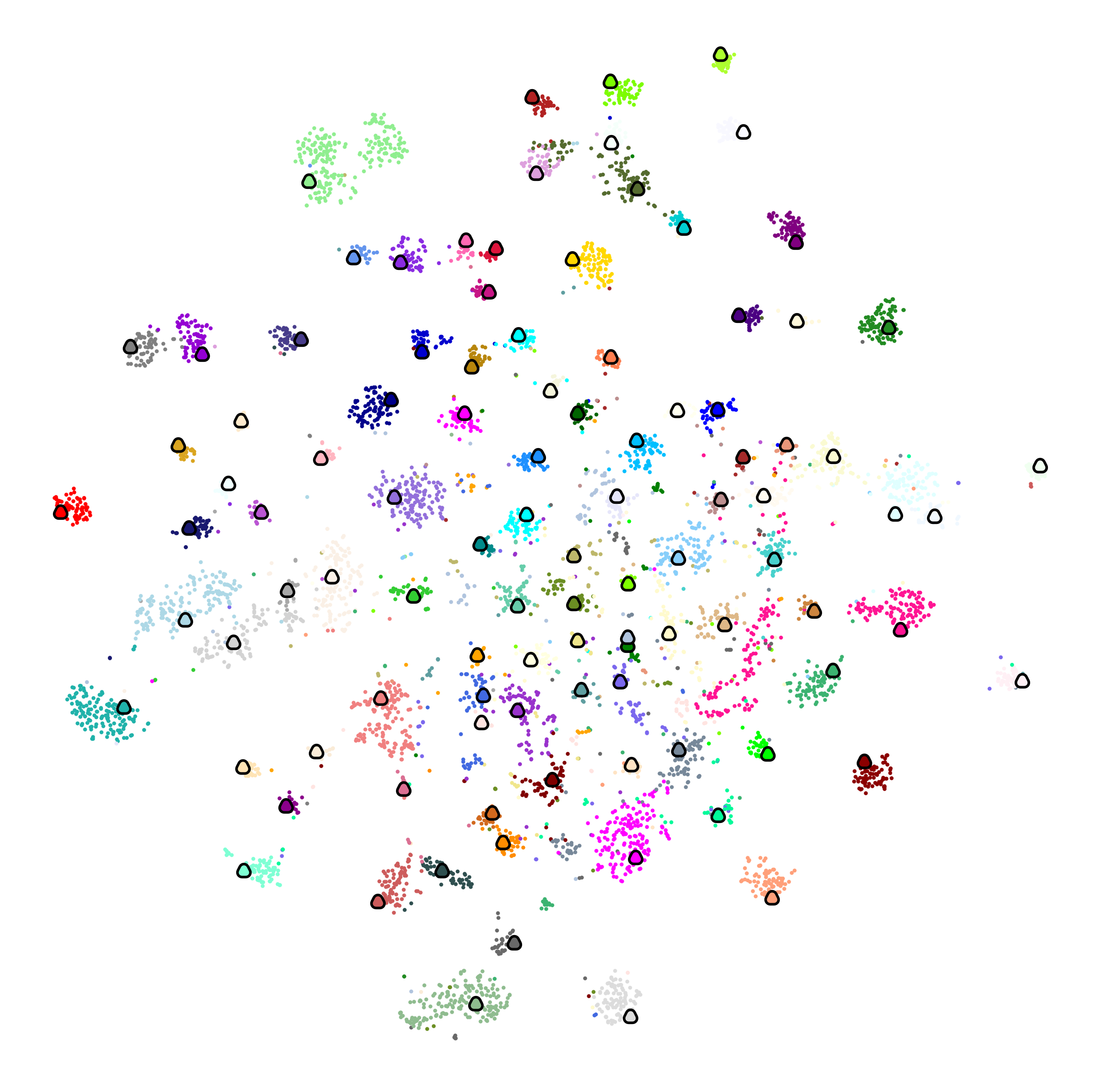}}
  \centerline{\footnotesize (c) Oxford Flowers102. }
\end{minipage}
\hfill
\begin{minipage}[b]{3cm}
  \centering
  \centerline{\includegraphics[width=3cm]{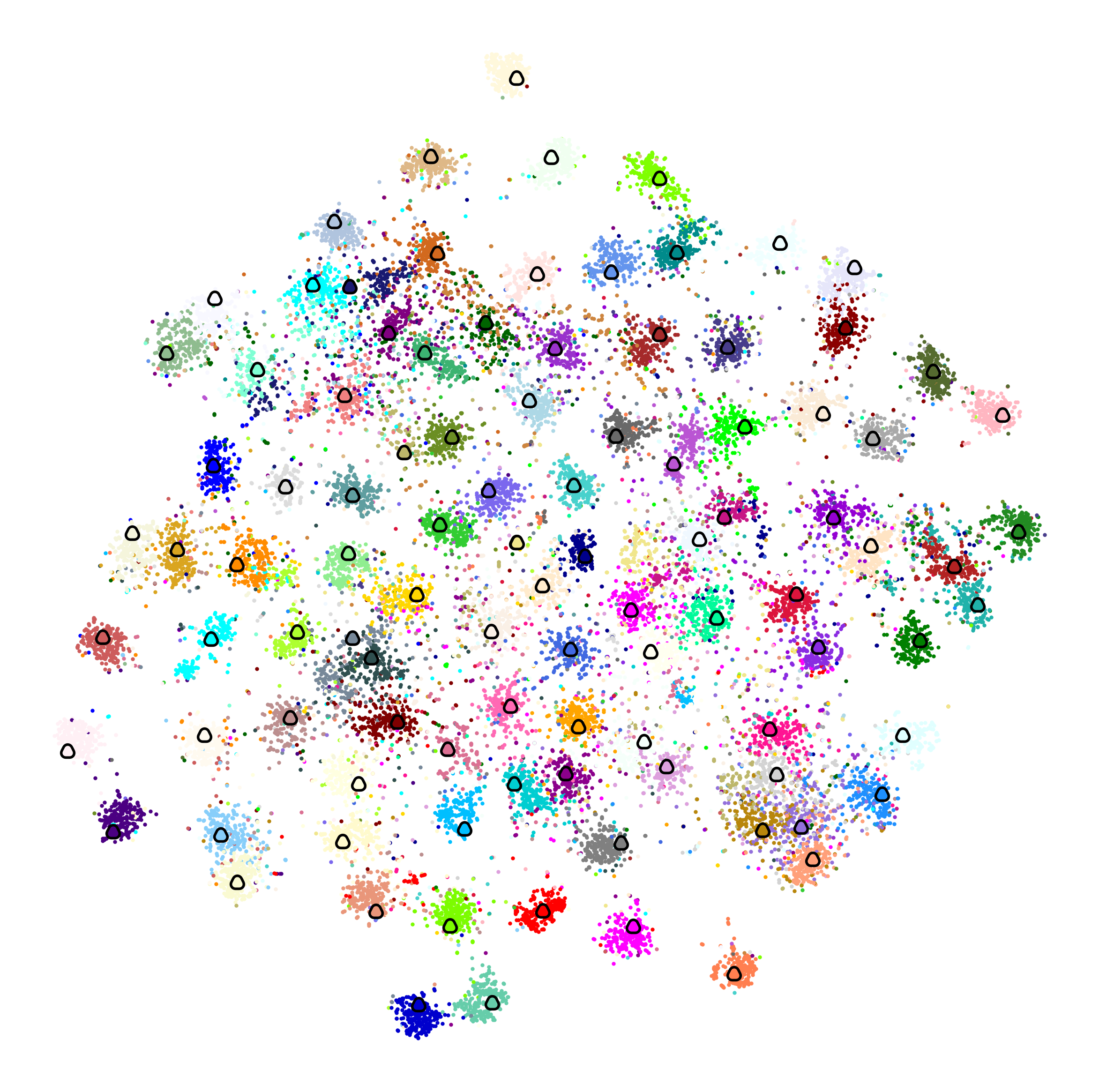}}
  \centerline{\footnotesize (d) Food101. }
\end{minipage}
\hfill
\begin{minipage}[b]{3cm}
  \centering
  \centerline{\includegraphics[width=3cm]{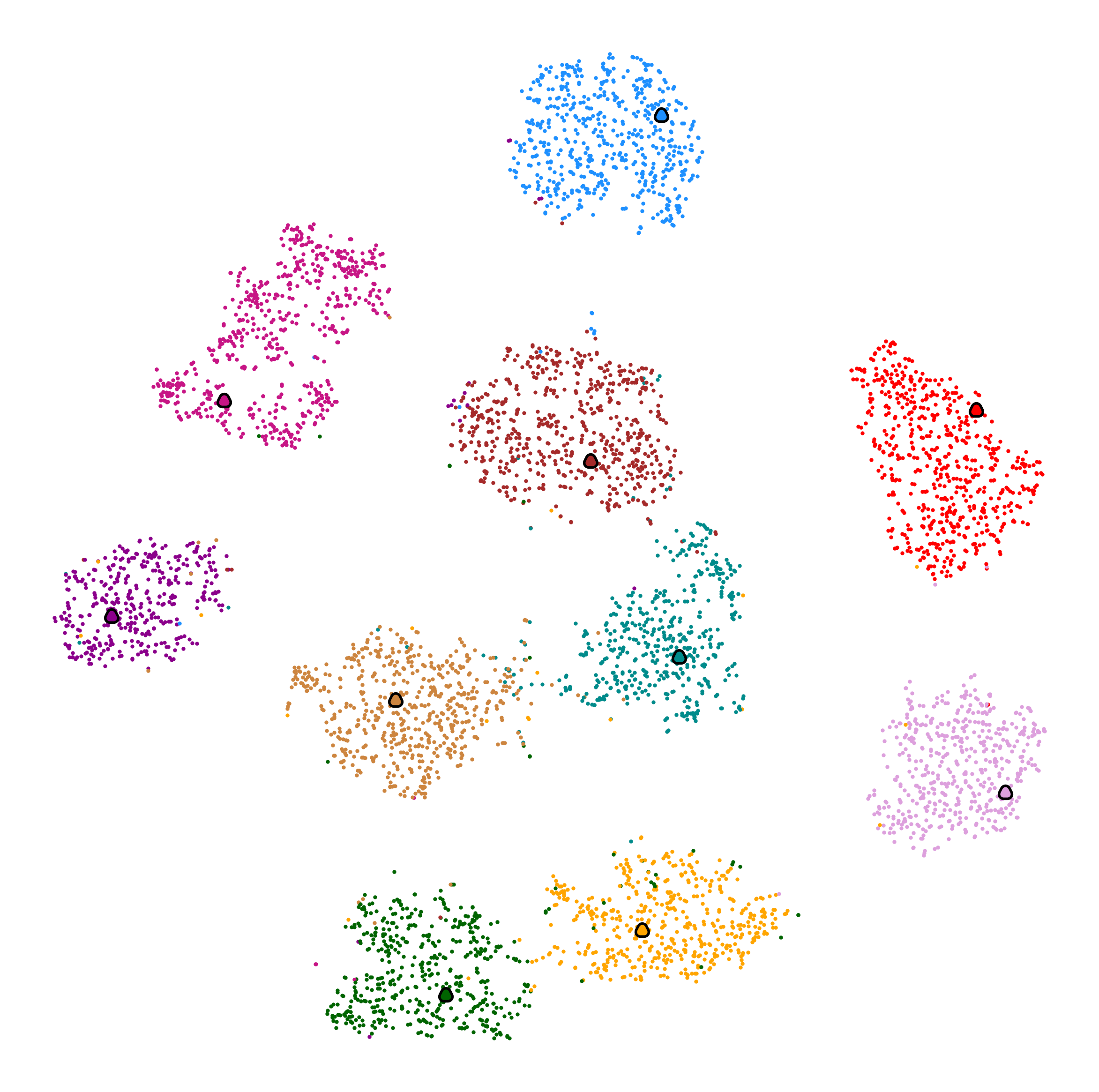}}
  \centerline{\footnotesize (e) EuroSAT.}
\end{minipage}

\begin{minipage}[b]{3cm}
  \centering
  \centerline{\includegraphics[width=3cm]{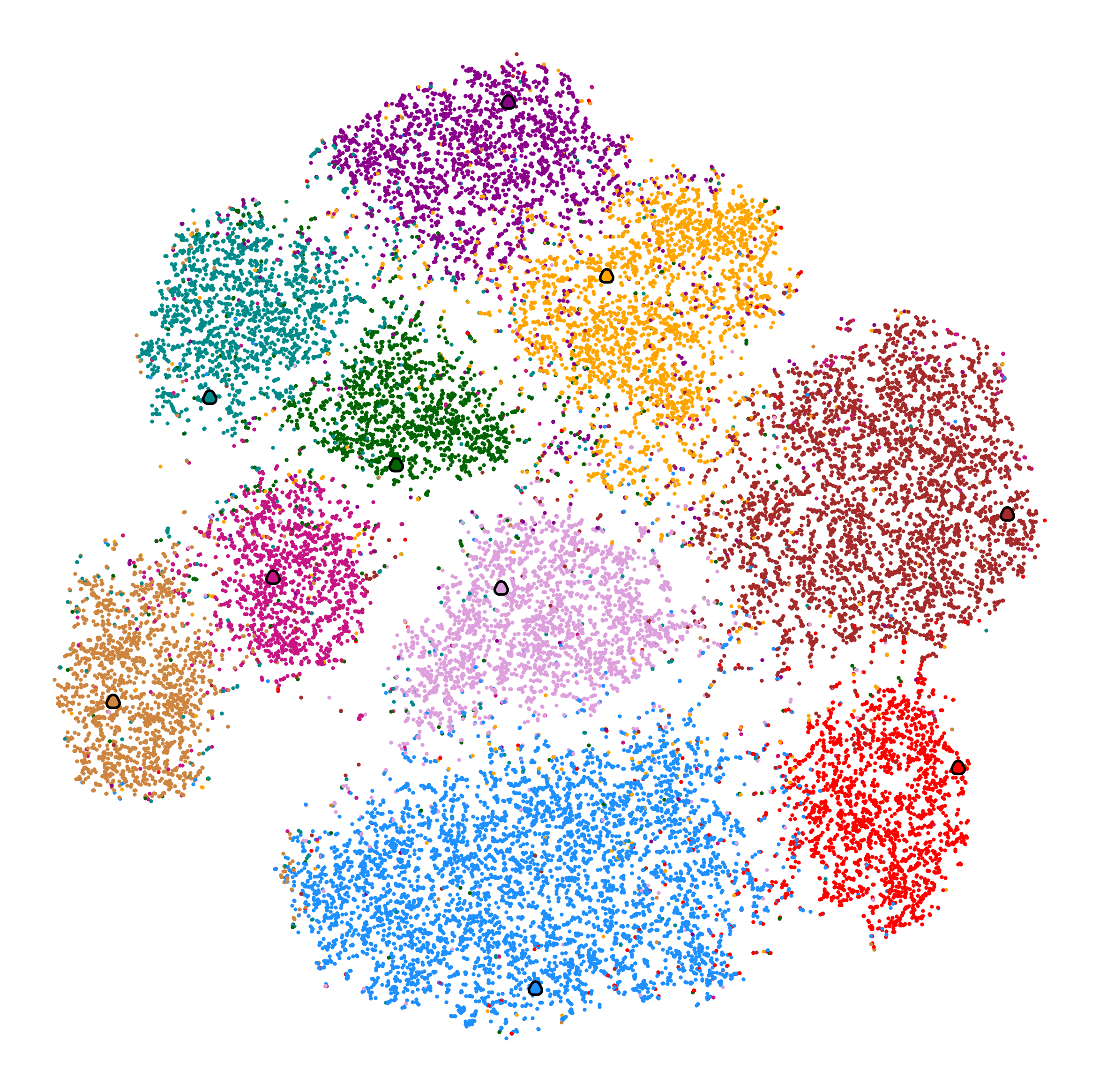}}
  \centerline{\footnotesize (f) SVHN.}
\end{minipage}
\hfill
\begin{minipage}[b]{3cm}
  \centering
  \centerline{\includegraphics[width=3cm]{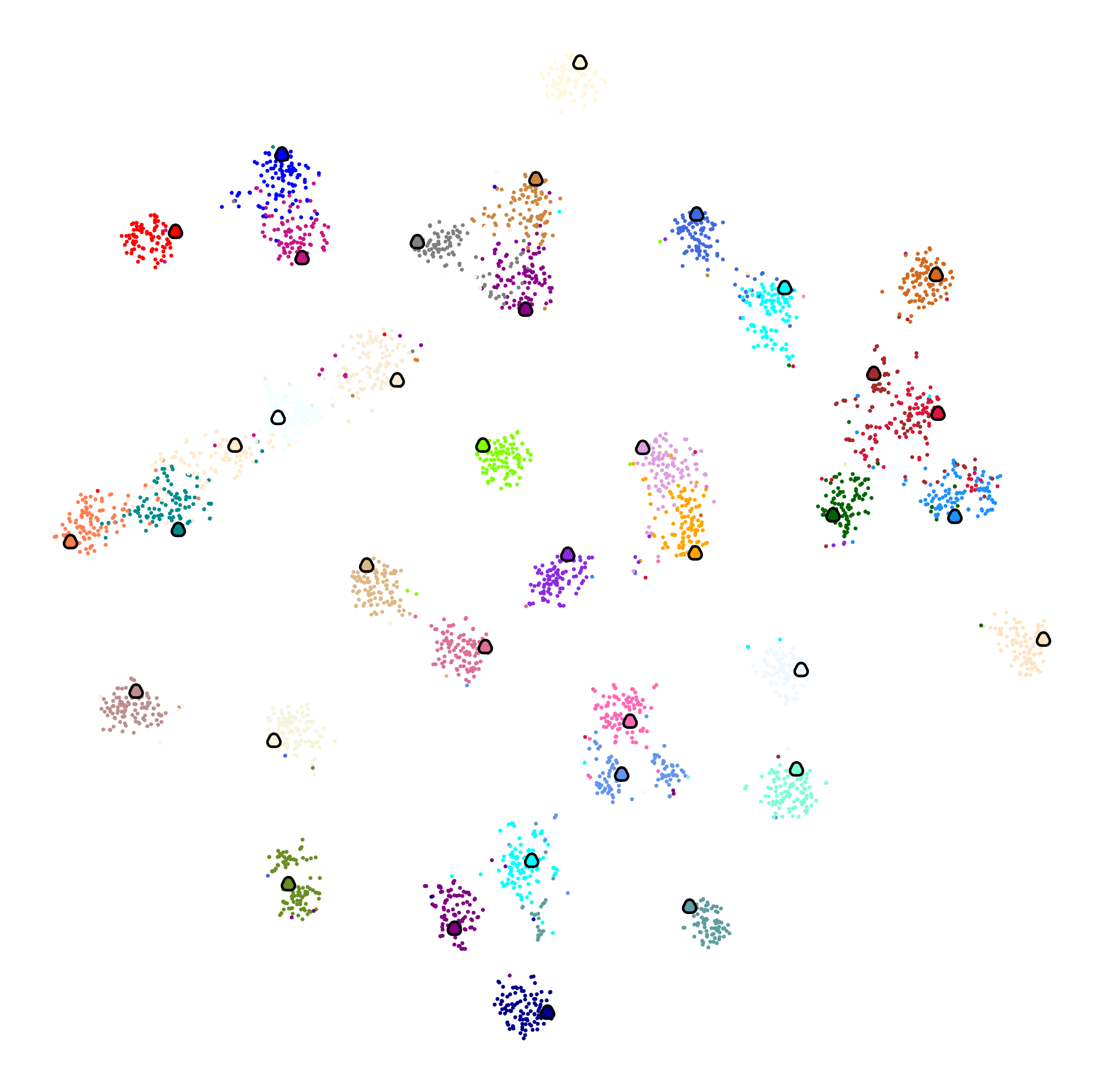}}
  \centerline{\footnotesize (g) Oxford Pets. }
\end{minipage}
\hfill
\begin{minipage}[b]{3cm}
  \centering
  \centerline{\includegraphics[width=3cm]{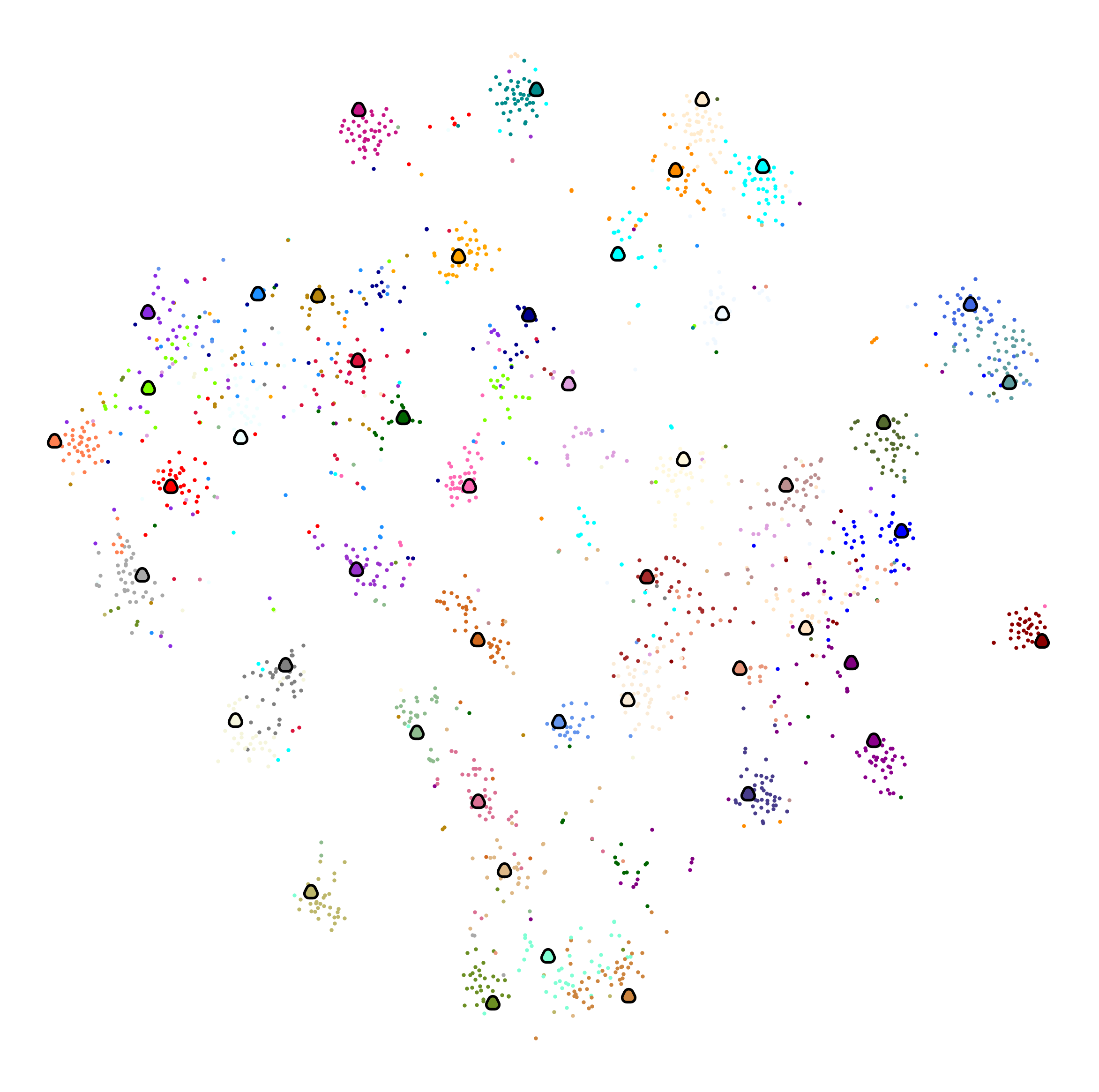}}
  \centerline{\footnotesize (h) DTD. }
\end{minipage}
\hfill
\begin{minipage}[b]{3cm}
  \centering
  \centerline{\includegraphics[width=3cm]{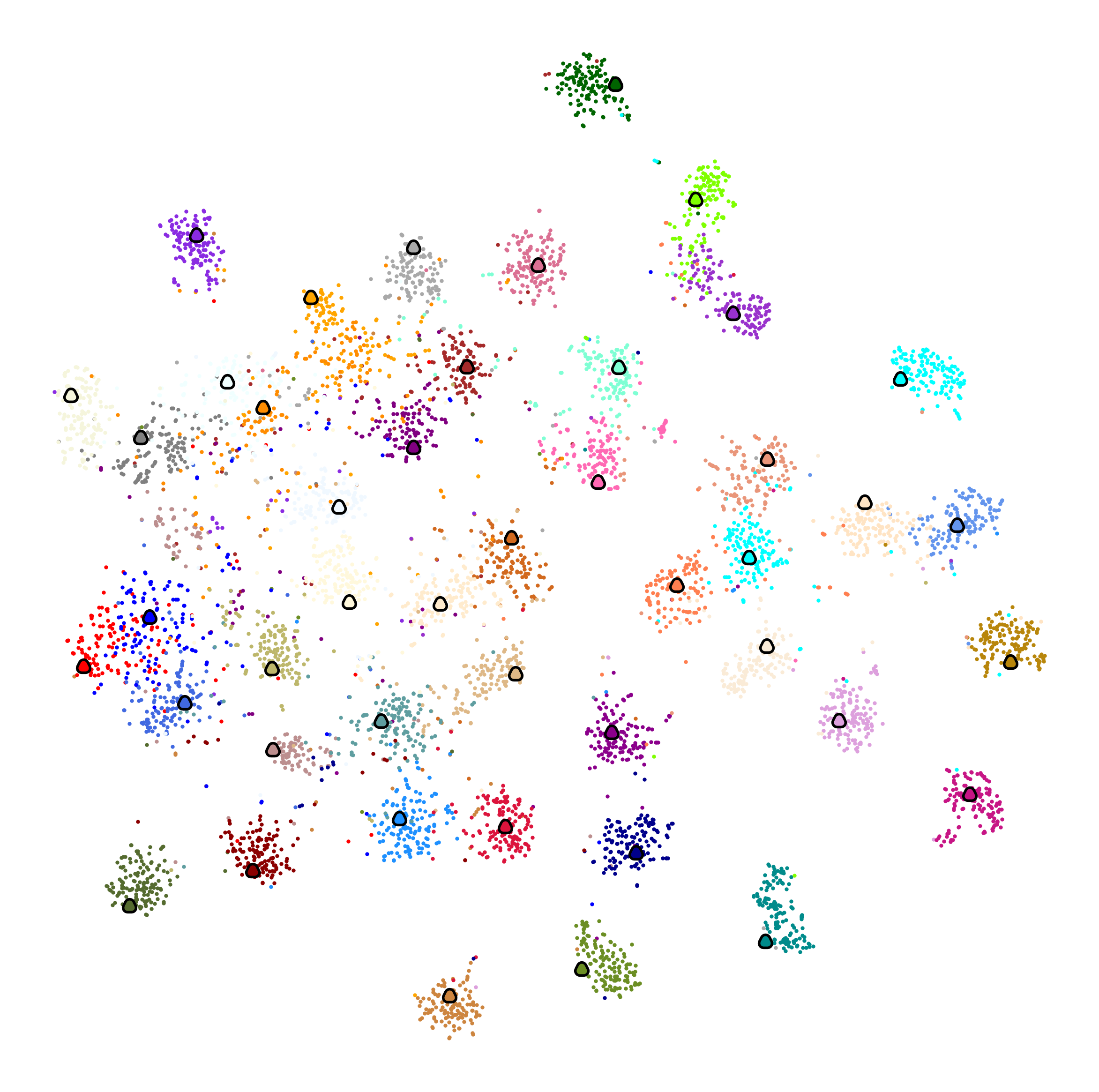}}
  \centerline{\footnotesize (i) Resisc45. }
\end{minipage}
\hfill
\begin{minipage}[b]{3cm}
  \centering
  \centerline{\includegraphics[width=3cm]{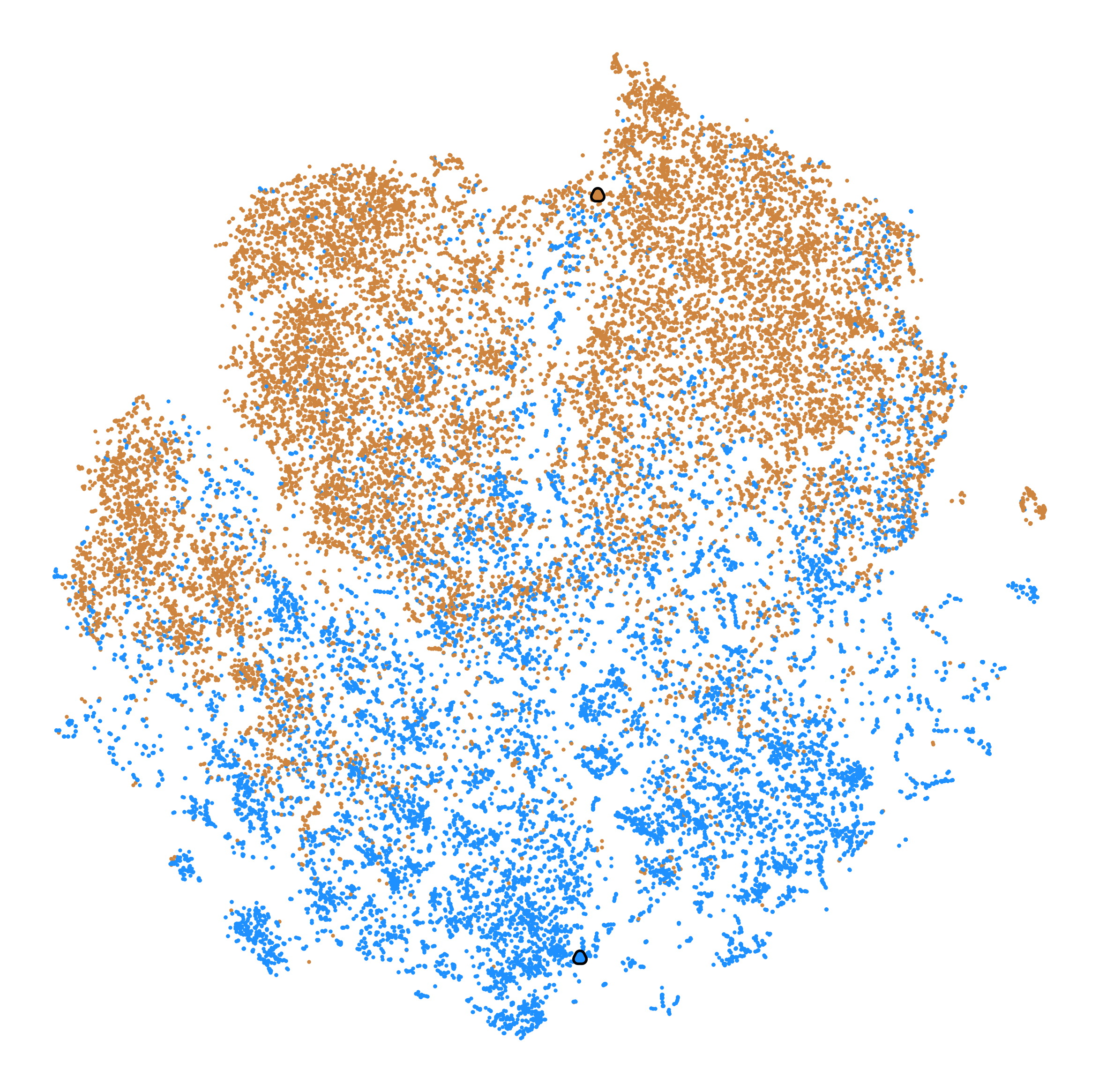}}
  \centerline{\footnotesize (j) Patch Camelyon. }
\end{minipage}
\hfill
\caption{Visualizations of the prototypes and the transformed tokens $\boldsymbol{u_{\tt [MASK]}}$ in testing phase using TSNE. Different colors represent different classes. The triangles with dark circles are the prototypes.} 
\label{fig:visualization}
\end{figure*}
To validate the parameter efficiency of VPTM, we compare our method with VPT under the optimal setting corresponding to Table~\ref{tab:compa_with_baseline} on prompt length $N_p$, GFLOPs, and the ratio of the amount of tuned parameters to the entire parameters $Tuned/Total$. Besides, the prototype vectors are also counted as optimized parameters in VPTM, we show the prototype dimension $t$ together in Table~\ref{tab:params}.

Regarding $N_p$, the average prompt length of VPTM is almost half of that of VPT. The average GFLOPs of VPTM is lower than that of VPT by $3.43$. Due to the existence of the parameters in verbalizer and prototypes, regarding the ratio $Tuned/Total$, the value of VPTM is $0.05\%$ higher than that of VPT. \emph{Though VPT tunes relatively fewer parameters than VPTM, VPTM is still more efficient from the GFLOPs comparison.} We analyze the reason as that the calculation cost is mostly caused by the token interaction. VPT requires two times of prompt tokens compared with our method, which results in more calculation burden. Therefore, by inheriting the pre-training task to keep consistency, VPTM is proved to be more efficient.

\subsection{Effectiveness Validation of the Prototypical Verbalizer}
To validate the effectiveness of the prototypical verbalizer, we replace it with the 1-layer and 2-layer MLP added on the prediction in the masked place to perform classification. The 1-layer MLP (MLP-1) directly maps the prediction in the masked place to the number of classes. The 2-layer MLP (MLP-2) firstly maps the prediction in the masked place to a 128 dimensional vector, which is then mapped to the number of classes. Accuracy and the ratio of the amount of tuned parameters to the entire parameters $Tuned/Total$ are delivered in Table.~\ref{tab:verba_compa}.

Compared with using the prototypical verbalizer, when using MLP-1, the amount of optimized parameters is fewer, and the average accuracy is lower by $8.18\%$. When increasing the amount of optimized parameters to be the same as that when using the prototypical verbalizer, the average accuracy achieved by using MLP-2 is still lower by $2.55\%$. The results demonstrate that mapping the prediction in the masked place to the classes is inferior to the prototypical verbalizer. In the pre-training phase, the {\tt [MASK]} token is supervised by the visual token. By inheriting the pre-training task, in our method, the prediction in the masked place is supposed to be like a word in the language vocabulary, but not equipped with explicit semantic meaning. It is not a comprehensive representation as the {\tt [CLS]} token. To achieve classification in this method, the rational way is to search for a mapping between the predicted token and downstream labels, but not to regard the {\tt [MASK]} as a comprehensive representation and simply conduct classification on it by adding MLPs.

In addition, to see the details of the prototypical verbalizer in constructing the mapping from predictions in the masked place to downstream labels, we visualize the distributions of the prototypes $\boldsymbol{c}_k\in\mathbb{R}^{t} , k \in [1,N_C]$ and the transformed tokens $\boldsymbol{u_{\tt [MASK]}}$ by TSNE, as shown in Fig.~\ref{fig:visualization}. For datasets that exhibit high accuracy, such as CIFAR10, EuroSAT and Resisc45, the transformed tokens $\boldsymbol{u_{\tt [MASK]}}$ predicted from the testing samples distribute tightly with their corresponding prototypes. For datasets on which the accuracy is not so high, including CIFAR100 and DTD, the prototypes can be separated clearly. There exists some overlap on the transformed tokens from different classes. We infer the reason as the low granularity of the visual tokens in the codebook.

\subsection{Ablation on Position of Prompts and {\tt[MASK]} Token }
\label{sec:ablate_pos}
\begin{figure}[tb!]
  \centering
  \includegraphics[width=7.5cm]{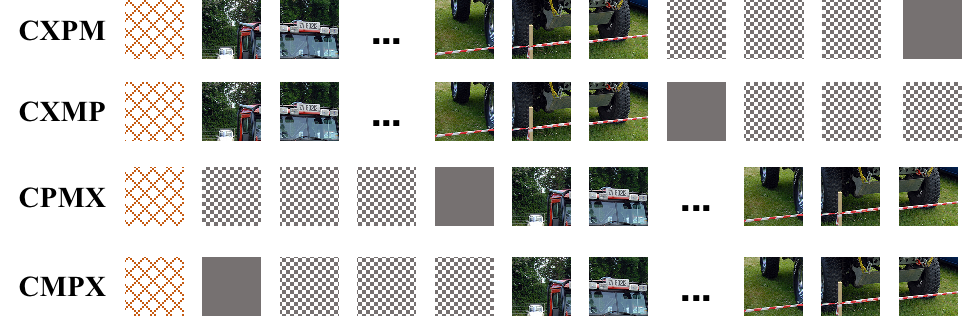}
  \caption{Four sets of positions that prompts and {\tt[MASK]} token relative to the original input. ``C": CLS token; ``X": image patches; ``P": prompt tokens; ``M": mask.}
  \label{fig:position}
\end{figure}

To show the impact of the position of prompts and {\tt[MASK]} token relative to the original input, we ablate 4 sets of positions that the prompts and {\tt[MASK]} token relative to the original input, as shown in Fig.~\ref{fig:position} The position relationships are represented by the order of their abbreviations. Based on the optimal setting in Table~\ref{tab:compa_with_baseline}, the ablation results are shown in Table~\ref{tab:position_ablation}. The greatest margin on the average accuracy is only $0.43$. We analyze the reasons of the stable performance as follows: 1) $40\%$ patches within each image are randomly block-wisely masked for pre-training, so that the pre-trained model can achieve relatively stable predictions regardless of the position of {\tt [MASK]} token; 2) more importantly, the VPTM inherits the pre-training task, bringing the robustness of VPTM against the positional changes.
\begin{table*}[t!]
  \centering
  \caption{The ablation study on the positions that the prompts and {\tt [MASK]} token relative to the original input. The order in the strings indicates the position relationships. ``C": {\tt[CLS]} token; ``X": image patch embeddings; ``P": prompts; ``M": {\tt[MASK]} token.}
  \footnotesize
  \label{tab:position_ablation}
  \begin{tabular}{l@{}| p{0.8cm}<{\centering} p{0.6cm}<{\centering} p{0.6cm}<{\centering} p{0.6cm}<{\centering} p{0.6cm}<{\centering} p{0.6cm}<{\centering} p{0.6cm}<{\centering} p{0.6cm}<{\centering} p{0.6cm}<{\centering} p{1.2cm}<{\centering}| p{1.0cm}<{\centering} }
    \toprule
    \midrule
    Positions\,\,\,\, & CIFAR100 & CIFAR10 & Flowers & Food101 & EuroSAT & SVHN & Pets & DTD & Resisc45 & PatchCame & Average\\
    \midrule
    CXPM  & 79.43 & 96.74 & 90.52 & 82.65 & 98.37 & 81.80 & 90.35 & 75.96 & 90.86 & 80.39 & 86.71 \\
    CXMP  & 73.77 & 96.90 & 90.58 & 81.84 & 97.30 & 90.43 & 90.62 & 75.27 & 89.75 & 81.29 & 86.78 \\
    CPMX  & 72.79 & 97.12 & 90.57 & 82.02 & 97.09 & 90.05 & 91.22 & 75.80 & 88.46 & 78.33 & 86.35 \\
    CMPX  & 72.55 & 96.95 & 90.67 & 82.64 & 97.09 & 91.35 & 91.41 & 75.59 & 88.63 & 79.65 & 86.65 \\
    \bottomrule
  \end{tabular}
\end{table*}

\subsection{Ablation on Prompt Length $N_p$}
\label{sec:ablate_promptLen}
\begin{figure}[tb!]
  \centering
  \includegraphics[width=7cm]{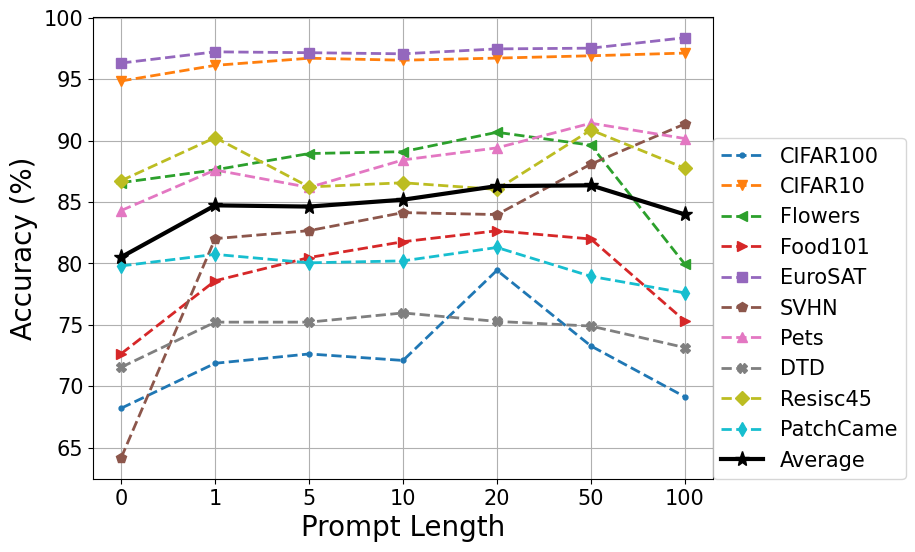}
  \caption{Accuracy under different settings of prompt length.}
  \label{fig:prompt_len}
\end{figure}

We compare the results under different prompt length $N_p \in [0, 1, 5, 10, 20, 50, 100]$. The results are shown in Fig.~\ref{fig:prompt_len}.

When $N_p=0$, which refers to only the verbalizer works, the lowest average accuracy $80.51$ is achieved. The results validate the necessity of introducing learnable prompts.

When $N_p>0$, most datasets such as EuroSAT and CIFAR10 are less likely to be impacted by the prompt length. Regarding the average accuracy as shown in the dark line in Fig.~\ref{fig:prompt_len}, the largest margin between the highest ($N_p=50, Acc=86.35$) and lowest ($N_p=100, Acc=83.98$) average accuracy is 2.37. Our method exhibits stable performance against the change of prompt length. Moreover, the average accuracy when $N_p=100$ (except $N_p=0$) is the lowest, while the results with shorter prompt length are even better. It indicates that our method does not rely on longer prompts, i.e., more parameters that can be optimized.

\subsection{Ablation on Prototype Dimension $t$}
\begin{figure}[tb!]
  \centering
  \includegraphics[width=7.5cm]{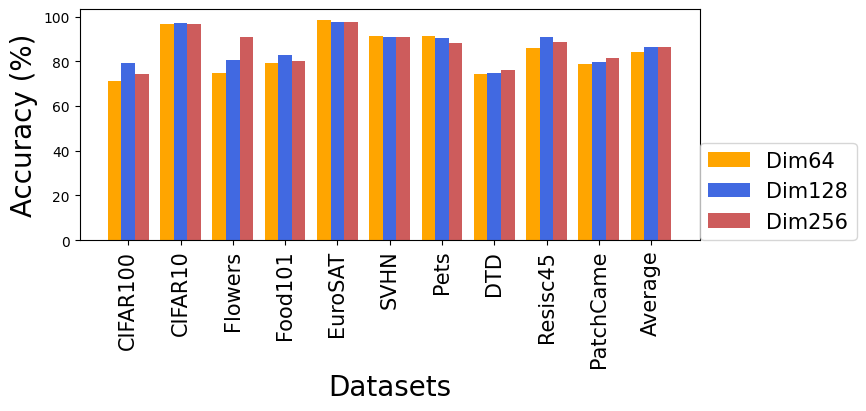}
  \caption{Accuracy under different prototype dimensions.}
  \label{fig:proto_dim}
\end{figure}

Under the setting of Table~\ref{tab:compa_with_baseline}, the comparisons on the dimension of prototypes $t \in [64, 128, 256]$ are given in Fig.~\ref{fig:proto_dim}. Almost equal performance is achieved on datasets such as CIFAR10 and SVHN. The highest average accuracy $86.41$ is achieved with $t=256$. Quite close to the highest one, the average accuracy when $t=128$ is $86.40$. When $t=64$, the average accuracy $84.26$ is the lowest. Our method also performs stably under different dimensions of prototypes. Considering the parameter-efficiency and the performance comprehensively, setting the dimension as $128$ is optimal.
\subsection{Significance of Semantics of Codebook}
\label{sec:sig_codebook}
To demonstrate the significance of the codebook's semantics, we compare the results of our method by using the weights of BEIT~\citep{bao2021beit} or BEITv2~\citep{peng2022beit}. BEIT~\citep{bao2021beit} has been pre-trained on ImageNet-21k~\citep{deng2009imagenet} with the codebook from DALL-E~\citep{ramesh2021zero}, while BEITv2~\citep{peng2022beit} has been pre-trained on ImageNet-1k~\citep{deng2009imagenet} with the codebook guided by CLIP~\citep{radford2021learning}. The results are given in Table~\ref{tab:compa_with_beitv1}.

Even though BEIT has been pre-trained on ImageNet-21k, the results of it are far worse than those of BEITv2, which has been pre-trained on ImageNet-1k. The codebook of pre-trained BEIT is from DALL-E, which enforces the model to recover low-level information of the image patches. In contrast, BEITv2 aims at exploring a semantic-rich visual tokenizer, which promotes the pre-training from learning the low-level pixel-wise features to the high-level semantic-wise features. The results validate the significance of the high-level semantics of the codebook.

\emph{Aiming at achieving the excellent performance of image classification by inheriting the masked visual token modeling pre-training, the visual tokens are supposed to be like the words in the vocabulary in NLP and carry high semantics.}

Taking a step further, there are only 8192 visual tokens in codebook in BEITv2, which are far fewer than words in vocabulary in NLP. To enrich the granularity of the visual tokens, pre-training with a larger codebook is required in the future.

\begin{table*}[t!]
  \centering
  \caption{The comparison of results of our method by using the weights of BEIT~\citep{bao2021beit} and BEITv2~\citep{peng2022beit}.}
  \footnotesize
  \label{tab:compa_with_beitv1}
  \begin{tabular}{l@{}| p{0.8cm}<{\centering} p{0.6cm}<{\centering} p{0.6cm}<{\centering} p{0.6cm}<{\centering} p{0.6cm}<{\centering} p{0.6cm}<{\centering} p{0.6cm}<{\centering} p{0.6cm}<{\centering} p{0.6cm}<{\centering} p{1.2cm}<{\centering}| p{1.0cm}<{\centering} }
    \toprule
    \midrule
    Methods & CIFAR100 & CIFAR10 & Flowers & Food101 & EuroSAT & SVHN & Pets & DTD & Resisc45 & PatchCame & Average\\
    \midrule
    VPTM + BEIT (ImageNet-21k)\,\,\,& 15.22 & 53.95 & 10.98 & 13.31 & 86.43 & 37.96 & \,7.93 & 22.23 & 57.27 & 74.92 & 38.02 \\
    VPTM + BEITv2 (ImageNet-1k)\,\,\, & 79.43 & 97.12 & 90.67 & 82.65 & 98.37 & 91.35 & 91.41 & 75.96 & 90.86 & 81.29 & 87.91 \\
    \bottomrule
  \end{tabular}
\end{table*}
\section{Discussions on the Backbone Dependence}
Following the core design of keeping consistency between downstream tasks and pre-training ones, VPTM is exactly suitable for BEITv2~\citep{peng2022beit}. Specifically, the large language models in NLP~\citep{kenton2019bert, zhang2019ernie} almost take the language modeling as pre-training task. Therefore, the \textit{cloze prompt} can be applied on language models that have been pre-trained by the \textit{masked language modeling} task, the \textit{prefix prompt} can be applied on language models that have been pre-trained by the \textit{casual (autoregressive) language modeling} task, for keeping the task consistency. In comparison, the pre-training tasks in vision area are various, e.g., supervised pre-training~\citep{dosovitskiy2020image}, masked image modeling (MIM)~\citep{he2022masked, chen2022context}, and masked visual token modeling (MVTM)~\citep{bao2021beit,peng2022beit}. The supervised pre-trained models are not equipped with generative task, and MIM pre-training task recovers each patch in pixel space, which lacks semantic-rich representations. They are not consistent with the cloze prompt. Thus, VPTM with a specific cloze prompt could not be applied on all pre-trained vision models.

On the other hand, the codebook also plays a key role, as analyzed in Section~\ref{sec:sig_codebook}. \emph{To achieve classification by mapping the visual tokens to downstream labels, visual tokens are supposed to be equipped with high-semantics, as the words in vocabulary in NLP}. Taking a step further, given the significant similarity between MLM and MVTM pre-training, also between cloze prompt in NLP and VPTM, it is expected to achieve unified multimodal prompting by inheriting the generative mask modeling pre-training in the vision-language area in the future.

\section{Conclusions}
In this paper, we propose the Visual Prompt learning as masked visual Token Modeling (VPTM), which is the first visual prompt method designed on generative pre-trained visual models and achieves consistency between pre-training and visual classification by task reformulation. Extensive experiments show that VPTM outperforms linear probe and CLIP-based visual prompt baselines. Compared with VPT, we also achieve the best average accuracy. The proposed VPTM is revealed to be parameter-efficient and easy to be deployed uniformly. Further ablation studies validate the effectiveness of the prototypical verbalizer, and exhibit the robustness of our method against the positions of prompts and {\tt [MASK]} token, prompt length and prototype dimensions. It demonstrates the rationality and efficacy of reformulating downstream tasks as the pre-training one to fulfill prompt learning in vision with task consistency.

\bmhead{Acknowledgments}
This work was partly supported at the university side, by National Science Foundation of China (62222607, 62002252), and Shanghai Municipal Science and Technology Major Project (2021SHZDZX0102).

\bmhead{Data Availability}
There is no new dataset included in this paper. The datasets in our experiments are all publicly available, they are: CIFAR100~\citep{krizhevsky2009learning}, CIFAR10~\citep{krizhevsky2009learning}, Oxford Flowers102~\citep{nilsback2008automated}, Food101~\citep{bossard2014food}, EuroSAT~\citep{helber2019eurosat}, SVHN~\citep{netzer2011reading}, Oxford Pets~\citep{parkhi2012cats}, DTD~\citep{cimpoi2014describing}, Resisc45~\citep{cheng2017remote}, Patch Camelyon~\citep{veeling2018rotation}.
\bibliography{sn-bibliography}% common bib file
% \input sn-article.bbl

\end{document}